\documentclass[runningheads]{llncs}
\usepackage{graphicx}

\usepackage{tikz}
\usepackage{comment}
\usepackage{adjustbox}
\usepackage{amsmath,amssymb} 
\usepackage{booktabs}
\usepackage{import}
\usepackage{makecell}
\usepackage{orcidlink}
\usepackage{tabularx}
\usepackage{xcolor}
\usepackage{soul}

\usepackage[title]{appendix}
\usepackage[font=small]{caption}
\usepackage[font=small]{subcaption}
\usepackage{hyperref}

\usepackage[accsupp]{axessibility}  

\begin{document}
\pagestyle{headings}
\mainmatter
\def\ECCVSubNumber{5140}  

\title{RealPatch: A Statistical Matching Framework for Model Patching with Real Samples} 

\titlerunning{RealPatch}

\author{Sara Romiti\inst{1}\orcidlink{0000-0002-1980-6798} \and
Christopher Inskip\inst{1}\orcidlink{0000-0002-3276-7498} \and
Viktoriia Sharmanska \inst{1,4}\orcidlink{0000-0003-0192-9308} \and 
Novi~Quadrianto \inst{1,2,3}\orcidlink{0000-0001-8819-306X}}
\authorrunning{S. Romiti et al.}

\institute{Predictive Analytics Lab (PAL), University of Sussex, United Kingdom \and 
BCAM Severo Ochoa Strategic Lab on Trustworthy Machine Learning, Spain \and 
Monash University, Indonesia \and Imperial College London\\
\email{\{s.romiti, c.inskip, sharmanska.v, n.quadrianto\}@sussex.ac.uk}}

\maketitle

\begin{abstract}
    Machine learning classifiers are typically trained to minimise the average error across a dataset. Unfortunately, in practice, this process often exploits spurious correlations caused by subgroup imbalance within the training data, resulting in high average performance but highly variable performance across subgroups.
	Recent work to address this problem proposes \textit{model patching} with CAMEL. This previous approach uses generative adversarial networks to perform intra-class inter-subgroup data augmentations, requiring (a) the training of a number of computationally expensive models and (b) sufficient quality of model's synthetic outputs for the given domain. 
	In this work, we propose RealPatch, a framework for simpler, faster, and more data-efficient data augmentation based on statistical matching. Our framework performs model patching by augmenting a dataset with real samples, mitigating the need to train generative models for the target task. 
	We demonstrate the effectiveness of RealPatch on three benchmark datasets, CelebA, Waterbirds and a subset of iWildCam, showing improvements in worst-case subgroup performance and in subgroup performance gap in binary classification. 
	Furthermore, we conduct experiments with the imSitu dataset with 211 classes, a setting where generative model-based patching such as CAMEL is impractical. We show that RealPatch can successfully eliminate dataset leakage while reducing model leakage and maintaining high utility. The code for RealPatch can be found at \url{https://github.com/wearepal/RealPatch}.
\end{abstract}

\section{Introduction}
\begin{figure}[t!]
    \centering
    \includegraphics[scale=0.75]{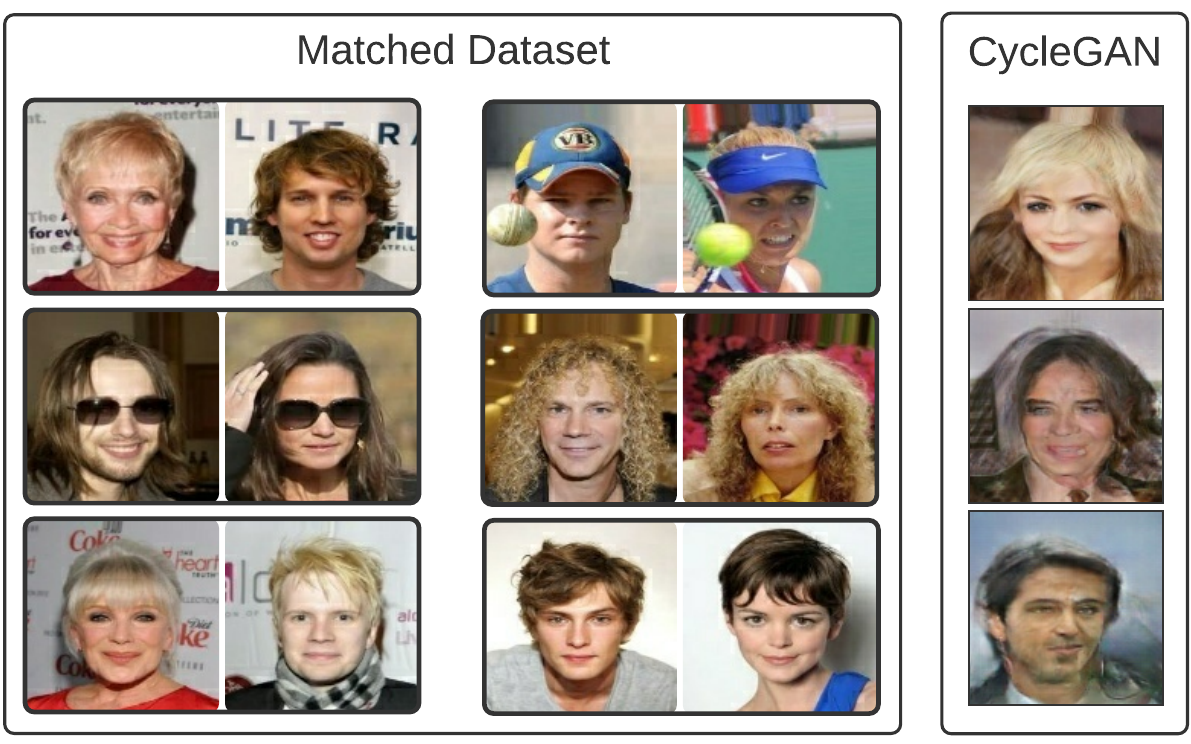}
    \caption{Examples of images and their counterfactuals on the attribute male/female, retrieved using RealPatch (left); both original and matched images are real samples from the CelebA dataset. RealPatch preserves characteristics across matched pairs such as pose, facial expression, and accessories. We also show CycleGAN synthetic counterfactual results (right) on the same attribute.}
    \label{fig:celeba_teaser}
\end{figure}
	
Machine learning models have fast become powerful yet ferocious pattern matching tools, able to exploit complex relationships and distant correlations present in a dataset. While often improving the average accuracy across a dataset, making use of spurious correlations (i.e. relationships that appear causal but in reality are not) for decision making is often undesirable, hurting generalization in the case that spurious correlations do not hold in the test distribution, and resulting in models that are biased towards certain subgroups or populations.  
	
Recent works in machine learning have studied the close link between invariance to spurious correlations and causation~\cite{PetBuhMei16,HeiMeiPet18,ArjBotGulPaz19,MitMcWWalHoletal21,VeiDamYadEis21}.
Causal analysis allows us to ask counterfactual questions in the context of machine learning predictions by relying on attribute-labelled data and imagining ``what would happen if"
some of these attributes were different. 
For example, ``would the prediction of a smile change had this person's cheeks been rosy''? 
While simple to answer with tabular data, generating counterfactuals for image data is non-trivial. 

Recent advances in generative adversarial networks (GAN) aim to build a realistic generative model of images that affords controlled manipulation of specific attributes, in effect generating image counterfactuals.
Leveraging research progress in GANs, several works now use counterfactuals for (a) detecting unintended algorithmic bias, e.g. checking whether a classifier's ``smile'' prediction flips when traversing different attributes such as ``heavy makeup'' \cite{DenHutMitGebetal20}, and (b) reducing the gap in subgroup performance, e.g. ensuring a ``blonde hair'' classifier performs equally well on male and female subgroups \cite{sharmanska2020contrastive,goel2020model}. The first relies on an \textit{invert then edit} methodology, in which images are first inverted into the latent space of a pre-trained GAN model for generating counterfactuals, while the latter uses an \textit{image-to-image translation} methodology. One of the most recent approaches, CAMEL \cite{goel2020model}, focuses on the latter usage of counterfactuals to \emph{patch} the classifier's dependence on subgroup-specific features.
	
GAN-based counterfactual results are encouraging, however, we should note that GAN models have a number of common issues such as mode collapse, failure to converge, and poor generated results in a setting with a large number of class labels and limited samples per class. We provide an alternative counterfactual-based model patching method that is simpler, faster, and more data-efficient. We focus on a statistical matching technique (see for example \cite{rubin1973matching}) such that for every image, we find an image with similar observable features yet having an opposite attribute value than the observed one; our counterfactual results are shown in Figure \ref{fig:celeba_teaser}. A statistical matching framework has been widely utilised to assess causality relationships in numerous fields, such as education \cite{morgan2001counterfactuals}, medical \cite{chastain1985estimated},\cite{christian2010prenatal},\cite{saunders1993association}, and community policies \cite{biglan2000value}, \cite{perry1996project} to name some. 
In this work, we explore statistical matching in the context of computer vision, and show its application for model patching with real samples.

Our paper provides the following contributions:
\begin{enumerate}
   \item We propose an image-based counterfactual approach for model patching called \emph{RealPatch} that uses real images instead of GAN generated images; 
    \item We provide an empirical evaluation 
    of different statistical matching strategies for vision datasets. Our results can be used as a \emph{guideline} for future statistical matching applications, for example showing the importance of using calipers;
	\item We show applications of RealPatch for improving the worst-case performance across subgroups and reducing the subgroup performance gap in a $2$-class classification setting. We observe that spurious correlation leads to shortcut learning, and show how RealPatch mitigates this by utilising a balanced dataset to regularise the training;
	\item We show applications of RealPatch for reducing dataset leakage and model leakage in a multi $211$-class classification setting. 
\end{enumerate}

\begin{figure*}[t!]
    \centering
    \includegraphics[scale=0.475]{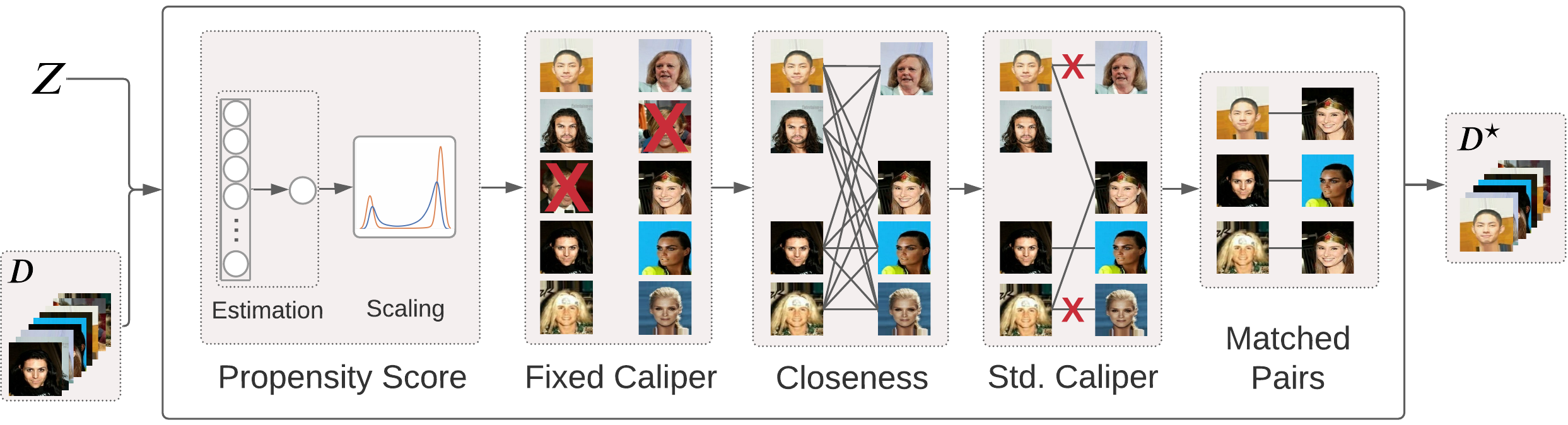}
    \caption{RealPatch: statistical matching pipeline. Given the dataset $D$ and the spurious attribute $Z$, the output is a matched dataset $D^{\star}$. To produce $D^{\star}$ we: 1) estimate the propensity score, and adjust it with temperature scaling; 2) restrict $D$ using the fixed caliper to remove \textit{extreme} samples; 3) compute the pair-wise closeness for each sample; 4) use the std-caliper to restrict the possible pairs according a maximum propensity score distance ; 5) for each sample, select the closest sample in the opposite group.}
    \label{fig:pipeline}
\end{figure*}
	
\textbf{Related Work.}
Data augmentation strategies using operations such as translation, rotation, flipping, cutout \cite{DeVTay17}, mixup \cite{ZhaCisDauLop18}, and cutmix \cite{YunHanOhChuetal19} are widely used for increasing the aggregate performance of machine learning models in computer
vision applications.
To improve performance in a targeted fashion, image transformation techniques that learn to produce semantic changes to an image are used to generate samples for underrepresented subgroups.
Sharmanska et al. \cite{sharmanska2020contrastive} used a StarGAN model \cite{ChoChoKimHaetal18} to augment the dataset with respect to a subgroup-specific feature, and subsequently optimized a standard Empirical Risk Minimization (ERM) training objective.
Whereas, CAMEL, a framework by Goel et al. \cite{goel2020model}, used a CycleGAN image transformation approach \cite{ZhuParIsoEfr17} and minimized a Sub-Group Distributionally Robust Optimization (SGDRO) objective function. 

GDRO method \cite{sagawa2019distributionally} aims to minimize the worst-case loss over groups in the training data.
CAMEL models \cite{goel2020model} minimize the class-conditional worst-case loss over groups.  
Another approach to reduce the effects of spurious correlation is optimizing a notion of invariance. 
Invariance serves as a proxy for causality, as features representing ``causes'' of class labels rather than ``effects'' will generalize well under intervention.
Invariant Risk Minimization (IRM) \cite{ArjBotGulPaz19} tries to find a data representation which discards the spurious correlations by enforcing that the classifier acting on that representation is simultaneously optimal in each subgroup. 
However, more analysis and better algorithms are needed to realize the promise of this framework in practice \cite{MitMcWWalHoletal21,goel2020model}.

Model patching focuses on robustness with respect to the unexpected failure of standard classifiers on subgroups of a class. 
Subgroups can correspond to environments/domains such as water or land, and can also refer to demographic attributes such as females or males \cite{CreJacZem21}.
Our work is therefore also related to many works addressing dataset biases in computer vision, particularly, in which the notion of bias relates to demographic attributes (e.g. \cite{Wang2019ICCV,YanQinFeiDenetal20,KehBarThoQua20,KehBarShaQua22}).
Wang et al. \cite{Wang2019ICCV} showed that even when datasets are balanced e.g. each class label co-occurs equally with each gender, learned models amplify the association between labels and gender, as much as if data had not been balanced. 
We refine their conclusions about balanced dataset and show that balancing with a statistical matching framework can successfully eliminate dataset leakage while reducing model leakage and maintaining high utility.

\section{Our RealPatch Framework}
We propose \textit{RealPatch}, a framework that first resamples a dataset such that the \textit{spurious groups} are balanced and equally informative and then utilise such dataset to regularise a classification objective.
RealPatch only uses real samples from the original dataset when constructing the augmented dataset, making it faster to perform, and simpler to apply to new tasks, compared to approaches such as CAMEL \cite{goel2020model} which require models to generate synthetic samples. 
Unlike standard data augmentation, our augmentation is in the context of statistical matching; it is a model-based approach for providing joint statistical information based on variables collected through two or more data sources.
If we have two sources, e.g. male and female, matching augments the source domain of male images with female images, and the domain of female images with male images.
In this section we outline the two stages of RealPatch. In Stage 1, a statistical matching procedure is used to construct a \emph{matched dataset}, a collection of comparable pairs of images that have opposite values of the spurious attribute. In Stage 2, we learn a model to predict the target label by including the representations of instances in the matched dataset.

\textbf{Setup.} Given a dataset of $N$ samples $D\!=\!\left\{1, \dots, N \right\}$ with target label $Y$ and spurious label $Z$, the dataset is divided into two \textit{spurious groups} $D_T$ and $D_C$ based on the value of $Z$. These partitions define the so-called \emph{treatment} ($Z\!=\!1$) and \emph{control} ($Z\!=\!0$) groups of size $N_T$ and $N_C$, respectively. Additionally, we call \textit{target groups} the two partitions created by $Y$ and \textit{subgroups} the four sets caused by both $Y$ and $Z$.
We use $X$ to denote feature representations of the input images extracted from a pre-trained model such as ResNet~\cite{he2016deep}, or Big Transfer BiT~\cite{kolesnikov2020big}.
In our framework these encoded representations $X$ are the \emph{observed covariates} that are used to compute the distances $M$ between images and identify the matched pairs. Following the work of causal inference, image representations in $X$ are assigned a propensity score, a measure of how likely an image $s$ belongs to the treatment group, $e_s = \hat{P}(Z_s\!=\!1|X_s)$. Propensity scores are used during Stage 1 to help prevent the inclusion of instances that would lead to poor matches. 
\subsection{Stage 1: Statistical Matching}\label{sec:matching}

Matching is a sampling method to reduce model dependency and enforce covariate balance in observational studies across a treatment and control group. 
In this work, we study the nearest-neighbour (NN) matching algorithm, which for each treatment sample selects the closest control sample. 
Figure~\ref{fig:pipeline} depicts our proposed matching pipeline. 
The pipeline has the following main building blocks: 1) \emph{propensity score estimation}; 2) \emph{closeness measure}; and 3) \emph{calipers} as a threshold mechanism. Before using the matched dataset in Stage 2, the \emph{matching quality} is measured by assessing the achieved balance of the covariates.

{\bf Propensity Score Estimation.} In causal inference, a propensity score $e_s$ is the probability of a sample $s$ being in the \textit{treatment} group $D_T$, given its observed covariates $X_s$.
This conditional probability is usually unknown, therefore it has to be estimated. This is typically done using a logistic regression on the observed $X$ to predict the binary variable $Z$ \cite{cox1989analysis}. Logistic regression allows us to reweight samples when optimising the loss function. We explore the use of \emph{spurious reweighting}, where samples are weighted inversely proportional to the frequency of their spurious label $Z$; more details are provided in Appendix A.
	
The shape of the conditional distribution has the potential to impact finding a suitable threshold. In this work we explore the use of \emph{temperature scaling} as a post-processing step to adjust the propensity score distribution before its use for matching. Temperature scaling has become a common approach for re-calibrating models \cite{guo2017calibration}, but to the best of our knowledge has not been utilised in the context of statistical matching for causal inference. In binary classification cases such as ours, for each sample $s$ the logits $z_s$ are divided by a (learned or fixed) parameter $t$ before applying the sigmoid function:
\begin{equation*}
    z_s = \text{log} \left( \frac{e_s}{1-e_s}\right), \ q_s = \frac{1}{1 + e^{-z_s / t}}. 
\end{equation*}
With $t\!=\!1$ we obtain the original probabilities. When $t\!<\!1$ the rescaled probabilities have a sharper distribution reaching a point mass at $t\!=\!0$. When $t\!>\!1$ the rescaled probabilities are smoother, reaching a uniform distribution as $t\! \rightarrow\!\infty$. As we show in our ablation study, we found rescaling to be beneficial for improving the achieved covariate balance (Table~\ref{tab:ablation_balance}).

{\bf Closeness Measure.} There are multiple metrics that can be used to measure the distance $M_{i,j}$ between samples $i \in D_T \text{ and } \: j \in D_C$, the most commonly used are \emph{Euclidean} and \emph{propensity score} distances.  
The \emph{Euclidean distance} is defined as $M_{ij} = (X_i - X_j)^{\top}(X_i - X_j)$ and the \emph{propensity score distance} as the distance between propensity scores $M_{ij} = |e_i - e_j|$.
Both \textit{Euclidean} and \textit{propensity score} distances have the advantage of being able to control how many samples are included via a threshold. While propensity score is the most commonly used matching method, Euclidean distance matching should be preferred \cite{king2019propensity} as the goal is to produce exact balance of the observed covariates rather than balance them on average.

{\bf Calipers.}	Nearest-neighbour matching is forced to find a match for every treatment sample and is therefore at risk of finding poor matched pairs. Caliper matching is a method designed to prevent matching samples with limited covariate overlap.
In this work we explore the usage of two different types of caliper, namely \emph{fixed caliper} and \emph{standard deviation (std) based caliper}, both applied to the estimated propensity score. \emph{Fixed caliper} \cite{crump2009dealing} is a selection rule that discards samples that have an estimated propensity score outside of a specific range; i.e. the dataset is restricted to $\left\{s, \: \forall \: s \in D \:|\: e_s \in [c, 1 - c]\right\}$. This allows the exclusion of examples with \textit{extreme} propensity scores; a rule-of-thumb used in previous studies \cite{crump2009dealing} considers the interval defined by $c\!=\!0.1$, i.e. $[0.1, 0.9]$. \emph{Standard deviation (std) based caliper} \cite{cochran1973controlling} is used to enforce a predetermined maximum discrepancy for each matching pair in terms of propensity score distance.
The distance $M_{ij}$ is kept unaltered if $|e_i - e_j| \le \sigma \cdot \alpha $, and is set to $\infty$ otherwise.
The variable $\sigma$ is the standard deviation of the estimated propensity score distribution and $\alpha$ is a parameter controlling the percentage of \textit{bias reduction} of the covariates. Cochran and Rubin \cite{cochran1973controlling} showed the smaller the $\alpha$ value the more the bias is reduced, the actual percentage of bias reduction depends on the initial standard deviation $\sigma$. Commonly used $\alpha$ values are $\left\{0.2, 0.4, 0.6\right\}$ \cite{cochran1973controlling}.

In our application we 1) restrict potential matches based on fixed caliper and 2) follow a hybrid approach selecting the closest sample using Euclidean distance matching while defining a maximum propensity score distance between samples. The final outcome of Stage 1 is a \emph{matched dataset} $D^{\star}$. 

{\bf Matching Quality.} Matching quality can be assesed through measuring the balance of the covariates across the treatment and control groups. Two commonly used evaluation measures are \emph{standardised mean differences (SMD)} and \emph{variance ratio (VR)} \cite{rubin2001using}. In the case that high imbalance is identified, Stage 1 should be iterated until an adequate level of balanced is achieved; we provide a guideline of adequacy for each metric below.
\emph{Standardised Mean Differences} computes the difference in covariate means between each group, divided by the standard deviation of each covariate. For a single covariate $\mathbf{a}$ from $X$ we have:\\
$$
\text{SMD} =   \frac{\bar{\mathbf{a}}_T- \bar{\mathbf{a}}_C}{\sigma} \text{ , where }\sigma=\sqrt{\frac{s^2_T + s^2_C}{2}}
$$
Here, $\bar{\mathbf{a}}_T$ ($\bar{\mathbf{a}}_C$) and $s^2_T $ ($s^2_C $) are respectively the sample mean and variance of covariate $\mathbf{a}$ in group $D_T$ ($D_C$). 
Intuitively, smaller \emph{SMD} values are better and as a rule of thumb an \emph{SMD} value below $0.1$ expresses an adequate balance, a value between $0.1$ and $0.2$ is considered not balanced but acceptable, and above $0.2$ shows a severe imbalance of the covariate~\cite{normand2001validating}. 
\emph{Variance Ratio} is defined as the ratio of covariate variances between the two groups, with an ideal value close to 1. While in some studies \cite{zhang2019balance} a variance in the interval $(0, 2)$ is defined acceptable, we follow Rubin \cite{rubin2001using} and use the stricter interval $(4/5, 5/4)$ to indicate the desired proximity to~1. To obtain a single measure for all covariates $X$, we categorise \emph{SMD} into $\le\!0.1$, $(0.1, 0.2)$, and $\ge\!0.2$, and \emph{VR} into $\le\!4/5$, $(4/5, 5/4)$, and $\ge\!5/4$ and assess the distribution of covariates. We show an assessment of matching quality for one run on each dataset in Section \ref{sec:ablation}, comparing the covariate balance before and after matching as well the effect of using temperature scaling.

\subsection{Stage 2: Target Prediction}\label{sec:target_prediction}
This stage is concerned with predicting a discrete target label $Y$ from covariates $X$. Inspired by Goel et al. \cite{goel2020model} our training process involves the minimization of a loss $\mathcal{L}$ that combines a SGDRO objective function $\mathcal{L}_{SGDRO}$ and a self-consistency regularisation term $\mathcal{L}_{SC}$:
\begin{equation}
    \mathcal{L} = \mathcal{L}_{SGDRO} + \lambda \mathcal{L}_{SC},
\end{equation}
where $\lambda$ is a hyperparameters controlling the regularisation strength.
The SGDRO loss is inspired by GDRO \cite{sagawa2019distributionally}, with the difference of considering a non-flat structure between the \textit{target} and \textit{spurious} labels; the hierarchy between target and spurious labels is included by considering the \textit{spurious groups} difference within each \textit{target group}. The SGDRO component of our loss is computed on the entire dataset $D$.

Similarly to \cite{goel2020model}, our $\mathcal{L}_{SC}$ encourages predictions $f_{\theta}(\cdot)$ of a matched pair $(x_T, x_C)$ in $D^{\star}$ to be consistent with each other and is defined as:
\begin{equation}
    \mathcal{L}_{SC}(x_T, x_C, \theta) = \frac{1}{2} \left[KL(f_{\theta}(x_T) || \tilde{m}) + KL(f_{\theta}(x_C) || \tilde{m})\right], 
\end{equation}
where $\tilde{m}$ is the average output distribution of the matched pair. 
While the SGDRO objective accounts for the worst-case subgroup performance, the form of the regularisation term induces  model's predictions to be subgroup invariant \cite{goel2020model}.

\section{Experiments}
We conduct two sets of experiments to assess the ability of RealPatch to 1) improve the worst-case subgroup performance and reduce the subgroup performance gap in a binary classification setting, and 2) reduce dataset and model leakage w.r.t. a spurious attribute in a 211-class classification setting. We describe them in turn.

\subsection{Reducing Subgroup Performance Gap}\label{sec:experiments_gap}	

In this section we study the effect of our RealPatch for increasing the worst-case performance across subgroups and reducing the gap in subgroup performance. We evaluate RealPatch against a variety baselines on three datasets, and perform an ablation analysis on configurations of RealPatch. We compare approaches using \textbf{Robust~Accuracy}: the lowest accuracy across the four subgroups, \textbf{Robust~Gap}: the maximum accuracy distance between the subgroups,
as well as \textbf{Aggregate~Accuracy}: a standard measure of accuracy. Our goal is to improve the robust accuracy and gap while retaining the aggregate accuracy performance as much as possible. That is because performance degradation on a subgroup(s) might occur if this improves the worst performing subgroup (e.g. \cite{martinez2020minimax}).

{\bf Datasets.} 
We use three publicly available datasets, CelebA\footnote{\scriptsize\url{http://mmlab.ie.cuhk.edu.hk/projects/CelebA.html}}~\cite{liu2015deep}, Waterbirds\footnote{\scriptsize\url{https://github.com/kohpangwei/group\_DRO}}~\cite{sagawa2019distributionally} and iWildCam-small\footnote{\scriptsize\url{https://github.com/visipedia/iwildcam_comp/tree/master/2020}}~\cite{beery2020iwildcam}. 
\textbf{CelebA} has 200K images of celebrity faces that come with annotations of 40 attributes. 
We follow the setup in \cite{goel2020model}, and consider hair colour $Y \in \left\{ \text{blonde}, \text{ non-blonde}\right\}$ as target label, and gender $Z \in \left\{ \text{male}, \text{ female}\right\}$ as spurious attribute. In this setup, the subgroup $\left(Y\!=\!\text{non-blonde}, Z\!=\!\text{female}\right)$ is under-sampled in the training set (from 71,629 to 4,054) as per \cite{goel2020model} amplifying a spurious correlation between the target and the demographic attribute. We keep all other subgroups as well as the validation and test sets unchanged. The images are aligned and resized to $128$x$128$. For stability we repeat our experiments three times using different randomly under-sampled subgroups $\left(Y\!=\!\text{non-blonde}, Z\!=\!\text{female}\right)$.
\textbf{Waterbirds} has 11,788 examples of birds living on land or in water. We follow \cite{sagawa2019distributionally} and predict $Y \in \left\{ \text{waterbird}, \text{ landbird}\right\}$, and use the background attribute $Z \in \left\{ \text{water}, \text{ land}\right\}$ as spurious feature. The spurious correlation between target and background is present in the dataset as waterbirds appear more frequently in a water scene, whereas landbirds on land. In order to perform three runs we randomly define the train/validation/test splits while enforcing the original subgroup sizes as per~\cite{goel2020model}.
\textbf{iWildCam-small} is a subset of iWildCam dataset~\cite{beery2020iwildcam}, whose task is to classify animal species in camera trap images. Here, we consider two species (meleagris ocellata and crax rubra) within two camera trap locations. The dataset contains $3,349$ images, specifically $2,005$ (train), $640$ (val) and $704$ (test). These splits have a spurious correlation between animal species and locations.This experiment emphasizes the applicability of RealPatch in a small dataset setting.

{\bf Baselines.}
Here we describe the four baseline methods used for comparison. \textbf{Empirical Risk Minimization (ERM)} is a standard stochastic gradient descent model trained to minimize the overall classification loss.  
\textbf{Group Distributionally Robust Optimisation (GDRO)} is a stochastic algorithm proposed by \cite{sagawa2019distributionally} with the aim of optimising the worst-case performance across the subgroups. \textbf{Sub-Group Distributionally Robust Optimisation (SGDRO)} \cite{goel2020model}
as described in Section \ref{sec:target_prediction}.
\textbf{CAMEL} is a two stage approach proposed by \cite{goel2020model} that uses the synthetic samples to define a subgroup consistency regulariser for model patching. Conceptually this model is most similar to ours, where we use real samples for model patching. 
The training details are in Appendix A.

\textbf{RealPatch Configurations and Hyperparameters.}
RealPatch can be instantiated in many different configurations. In these experiments hyperparameters of RealPatch include the choice of \textit{calipers}, \textit{temperature}, and \textit{reweighting strategies} in the propensity score estimation model as well as \textit{self-consistency strength} $\lambda$, \textit{adjustment coefficients} and \textit{learning rates} in the target prediction model. To select hyperparameters for Stage 1 of RealPatch we perform a grid search, selecting the configuration with the best covariates balance in term of \textit{SMD} and \textit{VR}. An ablation study on such hyperparameters is provided in Section~\ref{sec:ablation}. As per the hyperparameters of Stage 2, we perform model selection utilising the robust accuracy on the validation set. Further details of hyperparameters used and best configuration selected are summarised in Appendix A. 
    
\begin{table*}[t!]
    \begin{center}
    	\medskip
    	\caption{A comparison between RealPatch and four baselines on two benchmark datasets. The results shown are the average (standard deviation) performances over three runs. RealPatch is able to construct a model that is robust across subgroups with high robust accuracy and small robust gap.}
        \scalebox{.85}{
    		\begin{tabular}{lllll}
    			\toprule
    			\textbf{Dataset} & \textbf{Method} & \makecell{\textbf{Aggregate} $\uparrow$\\ \textbf{Accuracy (\%)}} & \makecell{\textbf{Robust} $\uparrow$\\ \textbf{Accuracy (\%)}} & \makecell{\textbf{Robust} $\downarrow$\\ \textbf{Gap (\%)}}\\
    			\midrule
    			\textbf{CelebA} & \textbf{ERM} & 89.21 (0.32) & 55.3 (0.65) & 43.48 (0.68)\\
    		    & \textbf{GDRO} & \textbf{90.47} (7.16) & 63.43 (18.99) & 34.77 (19.65)\\
    		    & \textbf{SGDRO} & 88.92 (0.18) & 82.96 (1.39) & 7.13 (1.67)\\
    			& \textbf{CAMEL} & 84.51 (5.59) & 81.48 (3.94) & \textbf{5.09} (0.44)\\   
    			& \textbf{RealPatch (Ours)} & 89.06 (0.13) & \textbf{84.82} (0.85) & 5.19 (0.9)\\
    			\midrule
    			\textbf{Waterbirds} & \textbf{ERM} & 86.36 (0.39) & 66.88 (3.76) & 32.57 (3.95)\\
    			& \textbf{GDRO} & \textbf{88.26} (0.55) & 81.03 (1.16) & 14.80 (1.15)\\
    			& \textbf{SGDRO} &  86.85 (1.71) & 83.11 (3.65) & 6.61 (6.01)\\
    			& \textbf{CAMEL} & 79.0 (14.24) & 76.82 (18.0) & 7.35 (5.66)\\
    			& \textbf{RealPatch (Ours)} & 86.89 (1.34) & \textbf{84.44} (2.53) & \textbf{4.43} (4.48)\\
    			
    			\bottomrule
    		\end{tabular}
    	}
    	\label{tab:results_aggregate}
    \end{center}
\end{table*}

\begin{table}[t!]
	\begin{center}
	\medskip
	\caption{Experiments with iWildCam-small\cite{beery2020iwildcam}. The results shown are the average (standard deviation) performances over three runs. CycleGAN-based CAMEL is not applicable for small training data (2K images).}
	\begin{adjustbox}{width=.7\textwidth}
	\begin{tabular}{llll}
	\toprule
	\textbf{Method} & \makecell{\textbf{Aggregate} $\uparrow$\\ \textbf{Accuracy (\%)}} & \makecell{\textbf{Robust} $\uparrow$\\ \textbf{Accuracy (\%)}} & \makecell{\textbf{Robust} $\downarrow$ \\ \textbf{Gap (\%)}} \\
	\midrule
    \textbf{ERM} & \textbf{79.97} (1.18) & 75.43 (3.01) & 19.65 (1.96) \\
	\textbf{SGDRO} & 78.55 (2.45) & 75.50 (3.58) & 14.28 (4.35) \\
	\textbf{RealPatch (Ours)} & 79.36 (2.09) & \textbf{76.70} (3.19) & \textbf{11.36} (4.87) \\
    \bottomrule
	\end{tabular}
    \end{adjustbox}
	\label{tab:results_iwildcam}
	\end{center}
\end{table}

\textbf{Results on CelebA.} 
From Table \ref{tab:results_aggregate}, RealPatch is able to significantly improve the worst-case subgroup performance and reduce the subgroup performance gap compared to the other baseline methods such as ERM, GDRO, SGDRO. Our proposed method improves the robust accuracy, robust gaps and aggregate accuracy with respect to the best baseline SGDRO by $1.86\%$, $1.94\%$ and $0.14\%$ respectively. When compared to CAMEL, RealPatch improves robust accuracy ($+3.34\%$), but slightly worsens the robust gap ($+0.1\%$). Compared with CAMEL, GDRO and SGDRO, RealPatch is very consistent across runs, with a standard deviation of $0.13$, $0.85$ and $0.9$ for aggregate accuracy, robust accuracy and robust gap, in contrast to $5.59$, $7.16$ and $0.18$ (aggregate accuracy), $3.94$, $18.99$ and $1.39$ (robust accuracy) and $0.44$, $19.65$ and $1.67$ (robust gap) for CAMEL, GDRO and SGDRO respectively.
%
On inspection of matched pairs from the dataset $D^{\star}$, we observe preservation in pose, facial expression (e.g. smiling in most of the examples), hair style (in many cases the colour as well, but not always), and accessories such as hat and glasses. Figures \ref{fig:celeba_teaser} shows samples of retrieved matched pairs, further examples are in Appendix B. Naturally, due to use of real samples used in matching, RealPatch suffers no issues regarding quality of images in the augmented dataset often observed with generative models. Figure \ref{fig:celeba_teaser} shows CycleGAN generated examples used in the consistency regularizer in the CAMEL's loss. 

\textbf{Results on Waterbirds.} Our RealPatch model can significantly reduce the gap between subgroup performances and improve the worst-case accuracy compared to all baselines. While GDRO have a better aggregate accuracy up to $1.37\%$, this model exhibits a higher imbalance over the subgroup performances with a robust gap of $14.80\%$ in comparison to $4.43\%$ of RealPatch and a robust accuracy of $81.03\%$ as opposed to $84.44\%$ of RealPatch. When compared to CAMEL, RealPatch shows improvements across all metrics, with $+7.89\%$ aggregate accuracy, $+7.62\%$ robust accuracy and $-2.92\%$ robust gap. Similar conclusions hold true when comparing RealPatch against the best baseline SGDRO with $+1.33\%$ robust accuracy and $-2.18\%$ robust gap.
The characteristics preserved between matched pairs are less obvious than in CelebA, mainly observing matching across the bird's primary/body colour; examples are shown in Appendix B. 

\textbf{Results on iWildCam-small.} We show results comparing RealPatch against ERM and the best baseline SGDRO in Table~\ref{tab:results_iwildcam}. When compared to the two baseline methods, our RealPatch improves robust accuracy by $+1.27\%$ and $+1.2\%$ and robust gap by $-8.29\%$ and $-2.92\%$.
It is worth noticing in this setting CycleGAN-based CAMEL is not applicable due to insufficient data to train CycleGAN. Examples of retrieved matched pairs are in Appendix B.

Please refer to Appendix B for the full results of Table~\ref{tab:results_aggregate} and Table~\ref{tab:results_iwildcam} which include the four subgroup performances.

\textbf{Ablation Analysis.}\label{sec:ablation}
\begin{figure*}[t!]
    \centering
    \includegraphics[scale=0.16]{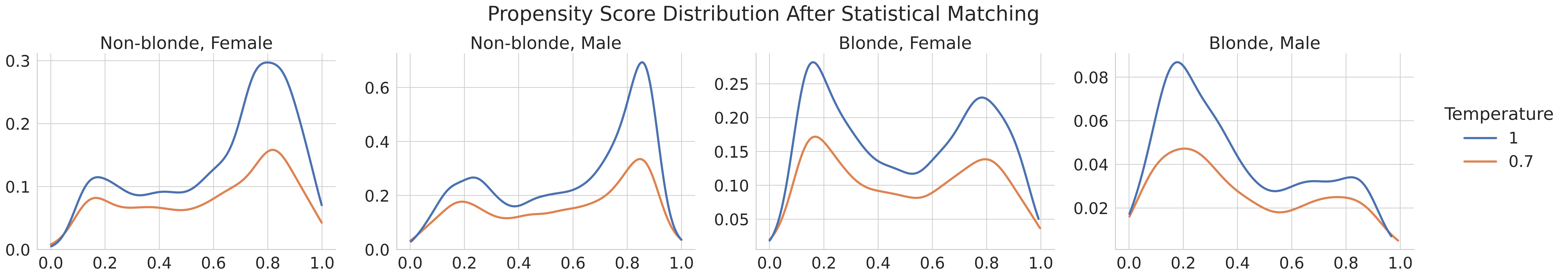}
    \caption{Estimated propensity score distribution on CelebaA dataset
     after matching, shown for each of the four subgroups. We compare the original distribution (blue, $t\!=\!1$) with its scaled version using the selected temperature (orange, $t\!=\!0.7$). Post-matching, the propensity score is approximately bimodal, showing that our procedure is balancing the propensity distribution across subgroups. Decreasing $t$ makes the two modes have more similar values, resulting in a matched dataset with better covariate balance in terms of \textit{SMD} and \textit{VR} (Table \ref{tab:ablation_balance}).}
    \label{fig:ablation_ps_celeba}
\end{figure*}
\begin{table}[t!]
\begin{center}
    \caption{Comparison of the covariate balance in 1) the original dataset $D$, 2) the matched dataset $D^{\star}$ 3) the matched dataset $D^{\star}$ with no temperature scaling 4) $D^{\star}$ with no fixed caliper and 5) $D^{\star}$ with no std-based caliper. The results are reported for a single run per dataset. Our matching procedure can successfully improve the covariate balance in both benchmark datasets, with fixed caliper significantly boosting its quality.}
	\medskip
	\scalebox{.85}{
	\begin{tabular}{llllllll}
		\toprule
		\textbf{Dataset} & & \multicolumn{3}{c}{\makecell{\textbf{SMD}}} & \multicolumn{3}{c}{\makecell{\textbf{VR}}} \\
        \midrule
		& & $\le 0.1$ $\uparrow$ & $(0.1, 0.2)$ $\downarrow$ & $\ge 0.2$ $\downarrow$ & $\le 4/5$ $\downarrow$ & $(4/5, 5/4)$ $\uparrow$ & $\ge 5/4$ $\downarrow$\\
		\cmidrule(lr){3-5} \cmidrule(lr){6-8}
		\textbf{CelebA} & $D$ & 348 & 344 & 1356 & 309 & 859 & 880\\ 
		& $D^{\star}$ (best) & \textbf{1977} & 71 & 0 & 0 & \textbf{2038} & 10 \\ 	
		& $D^{\star}\,\,(t\!=\!1)$ & 1957 & 91 & 0 & 2 & 2032 & 14 \\ 	
		& $D^{\star}\,\,(c\!=\!0)$ & 1522 & 482 & 44 & 13 & 1797 & 238 \\
		& $D^{\star}\,\,(\alpha\!=\!\infty)$ & 1909 & 138 & 1 & 11 & 2028 & 9 \\
		\midrule
		\textbf{Waterbirds} & $D$ & 376 & 346 & 1326 & 992 & 723 & 333 \\ 
		& $D^{\star}$ (best) & \textbf{1436} & 512 & 100 & 18 & \textbf{1533} & 497 \\ 	
		& $D^{\star}\,\,(t\!=\!1)$ & 1409 & 526 & 113 & 13 & 1482 & 553 \\ 
		& $D^{\star}\,\,(c\!=\!0)$ & 852 & 596 & 600 & 50 & 1104 & 894\\
		& $D^{\star}\,\,(\alpha\!=\!\infty)$ & 1436 & 512 & 100 & 18 & 1533 & 497 \\
		\bottomrule
	\end{tabular}
	}
	\label{tab:ablation_balance}
\end{center}
\end{table}
We perform ablations on three of the main components of our statistical matching stage, analysing the effect of 1) temperature scaling, 2) fixed caliper and 3) std-based caliper towards matching quality.
Since the distribution of propensity scores is altered by temperature scaling, and the propensity score is used by both types of calipers to exclude possible matches, these components are fairly coupled. We compare results obtained using different matching configurations, starting with the best configuration optimised in term of covariates balance achieved and (a) removing temperature scaling (setting $t\!=\!1$), (b) removing the fixed caliper (setting $c\!=\!0$) and (c) removing the std-based caliper (setting $\alpha\!=\!\infty$). 
In Table~\ref{tab:ablation_balance} we report the covariate balance before ($D$) and after ($D^{\star}$) matching for a single run of CelebA and Waterbirds, under all the three settings. 
It should be noted that the selected best configuration for all three runs of Waterbirds do not include the usage of std-based caliper, there is therefor no difference between $D^{\star} \text{(best)}$ and $D^{\star} (\alpha\!=\!\infty)$ in Table~\ref{tab:ablation_balance}. 
A similar analysis for iWildCam-small is in Appendix B.
We evaluate the matching quality through Standardised Mean Difference (\textit{SMD}) and Variance Ratio (\textit{VR}) as described in Section \ref{sec:matching}. All datasets benefit from matching, resulting in a better covariate balance than the original dataset. In CelebA we are able to produce an adequate balance ($\textit{SMD} \le 0.1$ and $4/5 \le \textit{VR} \le 5/4$) for most of the covariates, 1977 out of 2048 for \textit{SMD} and 2038 out of 2048 for \textit{VR}.
For the Waterbirds dataset, we achieve a slightly less prominent balance, nevertheless, improvements with respect to the original training dataset are achieved. 
Across all datasets the strongest effect is obtained by removing the influence of the \textbf{fixed caliper}. Since the propensity score characterises how likely an image is to belong to a subgroup, preserving all images at the extremes before calculating possible pairs is seen to be highly detrimental. The impact of \textbf{temperature scaling} and \textbf{std. caliper} is weaker overall, or absent for Waterbids under the setting $\alpha=\infty$, though still worth investigating for the specific application. 

While the use of calipers is relatively well known in causal inference, temperature scaling is not commonly explored. We inspect the effect of its usage on the propensity score distribution.
For a single run of CelebA, in Figure~\ref{fig:ablation_ps_celeba} we show the estimated propensity score distribution for each of the four subgroups for $D^{\star}$ (the after matching dataset). Post-matching, the propensity score is approximately bimodal, showing that our procedure is balancing the propensity distribution across the subgroups. 
We show the distribution of $D^{\star}$ generated with $t\!=\!1$ (no temperature) and  $D^{\star}$ generated with $t\!=\!0.7$ (selected temperature).
Decreasing $t$ leads to the two modes having more similar values, resulting in matched dataset with better covariate balance in terms of \textit{SMD} and \textit{VR} (Table \ref{tab:ablation_balance}). We observe a similar effect on the Waterbirds dataset, shown in Appendix B.

\subsection{Reducing Dataset and Model Leakage}
In this section we study the effect of our RealPatch on dataset and model leakage.

{\bf Leakage.} We use dataset leakage and model leakage \cite{Wang2019ICCV} to measure dataset bias. Dataset leakage measures how much information the true labels leak about gender, and corresponds to the accuracy of predicting gender from the ground truth annotations. In model leakage, the model is being trained on the dataset, and we measure how much the predicted labels leak about gender. 

{\bf imSitu dataset.}
We use the imSitu dataset \cite{yatskar2016} of situation recognition, where we have images of 211 activities being performed by an agent (person). We follow the setting of prior works \cite{zhao2017,Wang2019ICCV} and study the activity bias with respect to a binarised gender of the agent. 
The dataset contains 24,301 images (training), 7,730 images (validation), 7,669 images (test). 

\textbf{Matching Results.} We performed matching on the training data, and include all matched pairs as a rebalanced dataset to analyse the leakage. On this dataset, all samples have been matched, and the dataset size after matching has been doubled. This is expected, given the dataset has 211 classes with $44-182$ samples per class, which is significantly less than in CelebA.
Doubling the size of the dataset does not mean we include every sample twice. 
Instead this should be seen as rebalancing/resampling the dataset based on how many times each sample has been matched.
For matching we use the features extracted with a pre-trained ResNet101 model. 
The selected hyperparameters are \textit{spurious reweighting} in propensity score estimation, a temperature of $t\!=\!0.6$, $c\!=\!0$ in the fixed caliper, and $\alpha\!=\!0.2$ in the std-based caliper. This configuration was selected based on the best covariates balanced achieved on the training set: we can reach an adequate balance with an \emph{SMD} value below 0.1 and \emph{VR} close to 1 for most of the 1024 covariates used, 992 and 1010 respectively compared to 327 and 510 of the original dataset. A table with all the covariates balance is in Appendix B.
  

\textbf{Leakage Results.} 
We follow the same architectures and training procedures as \cite{Wang2019ICCV} to measure the dataset and model leakage. 
We compare our results with the rebalancing strategies based on gender-label co-occurances proposed in \cite{Wang2019ICCV}. 
We report our findings in Table \ref{table:imsitu}.
\begin{table*}[t!]
    \begin{center}
    	\medskip
    	\caption{Matching-based rebalancing in imSitu achieves the best leakage-accuracy trade-off. It shows nearly no dataset leakage, leading to a reduction in model leakage while maintaining overall accuracy. This is in contrast to the co-occurrence-based rebalancing based on gender-label statistics (e.g. $\alpha\!=\!1$ \cite{Wang2019ICCV}), where a reduction in dataset leakage does not lead to reduction in model leakage in a meaningful way, and the overall accuracy drops.}
        \scalebox{.9}{
    		\begin{tabular}{lllcc}
    			\toprule
    			\textbf{Data} & \makecell{\textbf{Dataset} $\downarrow$\\ \textbf{leakage} $\lambda_D$ } & \makecell{\textbf{Model} $\downarrow$\\ \textbf{leakage} $\lambda_M$ } & mAP {$\uparrow$} & F1 $\uparrow$\\
    			\midrule
                original training data  &68.35 (0.16)  &76.79 (0.17) &  41.12 & 39.91\\
    			\midrule
                {balancing with} $\alpha=3$ \cite{Wang2019ICCV} & 68.11 (0.55) &75.79 (0.49) & 39.20 & 37.64\\
                {balancing with} $\alpha=2$ \cite{Wang2019ICCV} & 68.15 (0.32) &75.46 (0.32) &37.53 & 36.41\\
                {balancing with} $\alpha=1$ \cite{Wang2019ICCV} & {53.99} (0.69) &74.83 (0.34) & 34.63 & 33.94\\
                RealPatch (ours) &55.13 (0.76) &68.76 (0.69) & 38.74 & 38.13\\
    			\bottomrule
    		\end{tabular}
    	}
    	\label{table:imsitu}
    \end{center}
\end{table*}
The results clearly show that dataset rebalancing via matching helps to achieve the best trade-off between debiasing (the dataset and the model leakage), and performance (F1 and mAP scores). We achieve significant reduction in dataset leakage (nearly no leakage $55.13$ versus original $68.35$) and model leakage ($68.76$ versus $76.79$), while maintaining the accuracy of the model with mAP and F1 scores comparable to those achieved with the original training data.
This is in contrast to rebalancing based on co-occurrences of gender and activity labels  \cite{Wang2019ICCV}. 
In the case of a rebalanced dataset with $\alpha\!=\!1$ that achieves nearly no dataset leakage ($53.99$), the model trained on this dataset leaks similarly to the model trained on the original data ($74.83$ versus $76.79$), and has a significant drop in the overall performance. 
This suggests that statistical matching helps to reduce dataset leakage in a meaningful way as the model trained on the rebalanced dataset can reduce leakage as well.  

\section{Limitations and Intended Use}

While \textit{SMD} and \textit{VR} are valuable metrics to indicate the quality of the matched dataset, there is no rule-of-thumb for interpreting whether the covariates have been \textit{sufficiently} balanced. Supplementing \textit{SMD} and \textit{VR} with manual inspection of matched pairs and evaluating on a downstream task is still required. 
 
Additionally, RealPatch currently only handles binary spurious attributes, requiring additional work (such as \cite{lopez2017estimation}) to handle matching over multiple treatments. It is worth noticing that also the baselines considered, GDRO, SGDRO and CAMEL, have only been tested on a binary spurious attribute. 
We intend to explore the usage of RealPatch for non-binary spurious attributes in the future work. 
A natural extension would be  to use a One-vs-Rest approach for matching: for each sample find the closest sample having a different value of the spurious attribute.

\section{Conclusions}

We present RealPatch, a two-stage framework for model patching by utilising a dataset with real samples using statistical matching. We demonstrate the effectiveness of RealPatch on three benchmark datasets, CelebA, Waterbirds and iWildCam-small.
We show that RealPatch's Stage 1 is successfully balancing a dataset with respect to a spurious attribute and we effectively improve subgroup performances by including such matched dataset in the training objective of Stage 2. 
We also highlight the applicability of RealPatch in a small dataset setting experimenting with the so-called iWildCam-small.
Compared to CAMEL, a related approach that requires the training of multiple CycleGAN models, we see competitive reductions in the subgroup performance gap without depending on the ability to generate synthetic images. We also show the effectiveness of RealPatch for reducing dataset leakage and model leakage in a 211-class setting, where relying on generative model-based patching such as CAMEL is impractical. RealPatch can successfully eliminate dataset leakage while reducing model leakage and maintaining high utility. Our findings show the importance of selecting calipers to achieve a satisfactory covariates balance and serve as a guideline for future work on statistical matching on visual data. 
We encourage the use of RealPatch as a competitive baseline for strategic rebalancing and model patching, especially in the case where developing models for image generation is prohibitive or impractical.
\newline\\
\textbf{Acknowledgments.} This research was supported by a European Research Council (ERC) Starting Grant for the project "Bayesian Models and Algorithms for Fairness and Transparency", funded under the European Union's Horizon 2020 Framework Programme (grant agreement no. 851538). NQ is also supported by the Basque Government through the BERC 2018-2021 program and by Spanish Ministry of Sciences, Innovation and Universities: BCAM Severo Ochoa accreditation SEV-2017-0718.
\clearpage

%
%
\bibliographystyle{splncs04}
\bibliography{bibfile}

\appendix
\section{Setup}\label{appendix:setting}
This section details our experimental setup for reproducibility, including dataset information and training details for the various baselines and proposed RealPatch framework. The code is made available at \url{https://github.com/wearepal/RealPatch}.

\subsection{Dataset}
Following the setup used by Goel et al. \cite{goel2020model}, ~\ref{tab:appendix_data_size} summarises sizes of each subgroup in both CelebA and Waterbirds. For each dataset, subgroup sizes are kept consistent across the three runs. The same information is provided for iWildCam-small, where 26 and 255 are the IDs of the two camera trap locations considered.
\begin{table}[h!]
	\centering
	\caption{Number of train/validation/test set images in each dataset.}
	\scalebox{0.8}{
	\begin{tabular}{llllll}
		\toprule
		\textbf{Dataset} & \textbf{Split} & \multicolumn{4}{c}{\textbf{Subgroup Size}} \\
		\midrule
		&  & Non-Blonde & Non-Blonde  & Blonde & Blonde\\
		&  & Female & Male  & Female & Male\\
		\cmidrule{3-6} 
		\textbf{CelebA} & \textbf{train} & 4054 & 66874 & 22880 & 1387\\
	    & \textbf{validation} & 8535 & 8276 & 2874 & 182\\
		& \textbf{test} & 9767 & 7535 & 2480 & 180\\
		\midrule
		&  & Landbird & Landbird  & Waterbird & Waterbird\\
		&  & Land & Water & Land & Water\\
		\cmidrule{3-6} 
		\textbf{Waterbirds} & \textbf{train} &  3498 & 184 & 56 & 1057\\
		& \textbf{validation} & 467 & 466 & 133 & 133\\
		& \textbf{test} & 2255 & 2255 & 642 & 642\\
		\midrule
		&  & Meleagris Ocellata & Crax Rubra & Meleagris Ocellata & Crax Rubra\\
		&  & ID 26 & ID 255  & ID 26 & ID 255\\
		\cmidrule{3-6} 
		\textbf{iWildCam-small} & \textbf{train} & 35 & 940 & 980 & 50\\
		& \textbf{validation} & 80 & 80 & 80 & 400\\
		& \textbf{test} & 85 & 80 & 90 & 449\\
		\bottomrule
	\end{tabular}}
	\label{tab:appendix_data_size}
\end{table}

\subsection{Baseline Training Details}\label{appendix:baselines_training}
For CelebA and Waterbirds all four baselines use a fine-tuned ResNet50 architecture, pre-trained on ImageNet. For ERM, GDRO and CAMEL we follow the setup used in \cite{goel2020model}. For each baseline, the hyperparameters selected are summarised in Table~\ref{tab:baseline_hyperparameter_values}. For iWildCam-small we use features extracted with a pre-trained BiT model to train both ERM and SGDRO; for ERM we use a logistic regression model with regularisation $C\!=\!1$, L2-penalty, tolerance of $1e^{-12}$ and sample weight inversely proportional to its subgroup frequency. For SGDRO we perform model selection using the robust accuracy on the validation set. 
We consider the following hyperparameters sweep for this baseline. 
For the Waterbirds dataset, adjustment coefficient is in a range of $\{2, 3, 5, 7\}$, weight decay is in a range of $\{0.005, 0.01, 0.05\}$ and batch size is in a range of $\{64, 128, 256 \}$.
For the CelebA dataset, adjustment coefficient is in a range of $\{2, 3, 5\}$, weight decay is in a range of $\{0.005, 0.01\}$, and batch size is fixed to 64. For the iWildCam-small dataset, adjustment coefficient is in a range of $\{1, 2\}$, weight decay is fixed to $0.01$ and batch size is in a range of $\{64, 128\}$.
For all datasets, we trained SGDRO for 100 epochs. The selected hyperparameters for each of the three runs are summarised in Table \ref{tab:baseline_hyperparameter_values_sgdro}.

\begin{table}[h!]
	\centering
	\caption{The hyperparameters used for ERM, GDRO, and CAMEL baselines for CelabA and Waterbirds, following~\cite{goel2020model}.}
	\scalebox{0.8}{
	\begin{tabular}{llllllll}
		\toprule
		\textbf{Dataset} & \textbf{Method} & \multicolumn{6}{c}{\textbf{Hyperparameters}} \\
		\midrule
		& & \textbf{Epochs} & \textbf{Learning} & \textbf{Weight} & \textbf{Batch} & \textbf{GDRO} & \textbf{$\lambda$}\\
 	    & & & \textbf{Rate} & \textbf{Decay} & \textbf{Size} & \textbf{Adjustment} &\\
		\cmidrule{3-8} 
		\textbf{CelebA} & \textbf{ERM} & 50  & 0.00005 & 0.05 & 16 & - \\
		& \textbf{GDRO} &  50 & 0.0001 & 0.05 & 16 & 3 & - \\
		& \textbf{CAMEL} &  50 & 0.00005 & 0.05 & 16 & 3 & 5 \\
        \midrule    
	    \textbf{Waterbirds} & \textbf{ERM} & 500  & 0.001 & 0.001 & 16 & -  & - \\
		& \textbf{GDRO} &  500 & 0.00001 & 0.05 & 24 & 1 & -  \\
		& \textbf{CAMEL} &  500 & 0.0001 & 0.001 & 16 & 2 &  100 \\
		\bottomrule
	\end{tabular}}
	\label{tab:baseline_hyperparameter_values}
\end{table}

\begin{table}[h!]
	\centering
	\caption{The hyperparameters used for the SGDRO baseline for each of the three runs.}
	\scalebox{0.8}{
	\begin{tabular}{lllll}
		\toprule
		\textbf{Dataset} & \textbf{Run} & \multicolumn{3}{c}{\textbf{Hyperparameters}} \\
		\midrule
		& & \textbf{Weight} & \textbf{GDRO} & \textbf{Batch}\\
 	    & & \textbf{Decay} & \textbf{Adjustment} & \textbf{Size} \\
		\cmidrule{3-5} 
		\textbf{CelebA} & \textbf{1} & 0.005 & 5 & 64 \\
		& \textbf{2} & 0.005 & 5 & 64 \\
		& \textbf{3} & 0.005 & 5 & 64\\
        \midrule
	    \textbf{Waterbirds} & \textbf{1} & 0.01 & 7 & 64 \\
		& \textbf{2} & 0.05 & 5 & 64 \\
		& \textbf{3} & 0.005 & 2  & 256\\
        \midrule
	    \textbf{iWildCam-small} & \textbf{1} & 0.01 & 2 & 128 \\
		& \textbf{2} & 0.01 & 2 & 64 \\
		& \textbf{3} & 0.01 & 1 & 64\\
		\bottomrule
	\end{tabular}}
	\label{tab:baseline_hyperparameter_values_sgdro}
\end{table}

\subsection{RealPatch Training Details}\label{appendix:realpatch}
To give each image a chance of being included in the final matched dataset $D^{\star}$, we match in both directions, i.e. we consider both values of the spurious attribute to represent the treatment and control group in turn. The size of $D^{\star}$ can therefore be in the range $\lbrack0, 2N\rbrack$; $0$ in the extreme case where no image is paired and $2N$ in the case that all images are. 
For example, in CelebA we first use our pipeline (Figure~\ref{fig:pipeline}) to match \textit{male} to \textit{female} samples, we then apply it to match \textit{female} to \textit{male} samples (using the same configuration and hyperparameters). 

\paragraph{Reweighting strategy.} In our logistic regression models for predicting the propensity score we explore the use of no reweighting, as well as a \textit{spurious-reweighting} strategy. For each sample $s$, its weight $w_s$ is defined as: 
\begin{equation*}
    w_s = \frac{N}{2 \cdot N_{z_s}},
\end{equation*}
where $N_{z_s}$ is the size of the spurious group $\left(Z\!=\!z_s\right)$.


\paragraph{Hyperparameters for Reducing Subgroup Performance Gap.} We include the hyperparameter sweep and provide the best hyperparameters found for each dataset and run. To select the hyperparameters for Stage 1 of RealPatch we perform a grid search summarised in Table \ref{tab:realpatch_hyperparameter_search}, selecting the configuration with the best covariates balance in terms of \textit{SMD} and \textit{VR}. Although we need to perform hyperparameters search, we notice the optimal values (Table~\ref{tab:realpatch_hyperparameter_values}) are quite stable across different seeds; in practice, the grid search for Stage~1 can be restricted. As per the hyperparameters of Stage 2, we perform model selection utilising the robust accuracy on the validation set. 
We consider the following hyperparameters sweep. For the Waterbirds dataset, adjustment coefficient is in a range of $\{2, 3, 5, 7\}$, weight decay is in a range of $\{0.005, 0.01, 0.05\}$, regularisation strength $\lambda$ is in a range of $\{0, 1, 5, 10\}$ and batch size is in a range of $\{64, 128, 256 \}$. 
For the CelebA dataset, adjustment coefficient is in a range of $\{2, 3, 5\}$, weight decay is in a range of $\{0.005, 0.01\}$, $\lambda$ is in a range of $\{0, 1, 5\}$, and batch size fixed to 64.
For the iWildCam-small dataset, adjustment coefficient is in a range of $\{1, 2\}$, weight decay is fixed to $0.01$, $\lambda$ is in a range of $\{0, 1, 2, 7, 10, 12, 15\}$, and batch size is in a range of $\{64, 128\}$.
Table \ref{tab:realpatch_hyperparameter_values} reports the values of the best hyperparameters found.

\begin{table}[h!]
    \begin{center}
    \caption{Hyperparameter grid search used in Stage 1 of RealPatch for reducing subgroup performance gap.}
    \scalebox{0.8}{
    \begin{tabular}{ll}
    \toprule
    \textbf{Hyperparameter} & \textbf{Sweep} \\
    \midrule
    \textbf{PS-reweighting} & no reweighting\\ 
    & spurious reweighting \\ 
    \textbf{PS-temperature} (t) & $\lbrack0.6, 1.3\rbrack$ with step $0.05$\\
    \textbf{Fixed caliper} ($c$) & 0.1\\
    & 0.05\\
    & 0 (None)\\
    \textbf{Std-based caliper} ($\alpha$) & 0.2 \\
    & 0.4 \\
    & 0.6 \\
    & $\infty$ (None) \\
    \bottomrule
    \end{tabular}}
    \label{tab:realpatch_hyperparameter_search}
    \end{center}
\end{table}

\begin{table}[h!]
	\begin{center}
	\caption{The hyperparameters values selected for RealPatch on CelebA, Waterbirds and iWildCam-small across three runs.}
	\scalebox{0.8}{
		\medskip
		\begin{tabular}{lllllllll}
			\toprule
			\multicolumn{6}{c}{\textbf{CelebA} dataset} \\ 
			\midrule
			\textbf{Run} & \textbf{PS-reweighting} & $t$ & \textbf{$c$} & \textbf{$\alpha$} & \textbf{Weight} & \textbf{GDRO} & \textbf{Reg.} & \textbf{Batch} \\
			& & & & & \textbf{Decay} & \textbf{Adj.} & $\lambda$ & \textbf{Size} \\
			\cmidrule{2-9} 
			\textbf{1} & no reweighting & 0.7 & 0.1 & 0.6 & 0.01 & 5 & 5 & 64\\
			\textbf{2} & no reweighting & 0.7 & 0.1 & 0.6 & 0.005 & 5 & 1 & 64\\
			\textbf{3} & no reweighting & 0.7 & 0.1 & 0.6 & 0.005 & 5 & 1 & 64\\
			\midrule
			\multicolumn{6}{c}{\textbf{Waterbirds} dataset} \\ 
			\midrule
			\textbf{Run} & \textbf{PS-reweighting} & $t$ & \textbf{$c$} & \textbf{$\alpha$} & \textbf{Weight} & \textbf{GDRO} & \textbf{Reg.} & \textbf{Batch} \\
			& & & & & \textbf{Decay} & \textbf{Adj.} & $\lambda$ & \textbf{Size} \\
			\cmidrule{2-9} 
			\textbf{1} & no reweighting & 0.9 & 0.1 & $\infty$ & 0.05 & 3 & 1 & 128\\
			\textbf{2} & no reweighting & 0.9 & 0.1 & $\infty$ & 0.05 & 3 & 1 & 128\\
			\textbf{3} & no reweighting & 0.7 & 0.1 & $\infty$ & 0.005 & 2 & 1 & 256\\
			\midrule
			\multicolumn{6}{c}{\textbf{iWildCam-small} dataset} \\ 
			\midrule
			\textbf{Run} & \textbf{PS-reweighting} & $t$ & \textbf{$c$} & \textbf{$\alpha$} & \textbf{Weight} & \textbf{GDRO} & \textbf{Reg.} & \textbf{Batch} \\
			& & & & & \textbf{Decay} & \textbf{Adj.} & $\lambda$ & \textbf{Size} \\
			\cmidrule{2-9} 
			\textbf{1} & spurious-reweighting & 1 & 0.05 & $\infty$ & 0.01 & 2 & 5 & 128\\
			\textbf{2} & spurious-reweighting & 1.3 & 0.1 & $\infty$ & 0.01 & 1 & 12 & 128\\
			\textbf{3} & spurious-reweighting & 1 & 0.05 & $\infty$ & 0.001 & 1 & 10 & 64\\
			\bottomrule
		\end{tabular}}
	\label{tab:realpatch_hyperparameter_values}
	\end{center}
\end{table}

\paragraph{Hyperparameters for Reducing Dataset and Model Leakage.} For the imSitu dataset we perform a grid search over hyperparameters, using \textit{spurious reweighting} in the propensity score estimation model, temperature $t\!=\!\lbrack0.6, 1\rbrack$ with step $0.1$, a fixed caliper with $c\!=\!\left\{0, 0.1\right\}$, and an std-based caliper with $\alpha\!=\!0.2$. For model selection, we use the covariate balanced achieved on the training set in terms of \textit{SMD} and \textit{VR}. The selected hyperparameters are \textit{spurious reweighting}, $t\!=\!0.6$, $c\!=\!0$, and $\alpha\!=\!0.2$.

\section{Results}
In Appendix \ref{sec:realpacth_results} we show additional results for our RealPatch framework. In Appendix \ref{sec:camel_results} we report the results obtained using different setups for the CAMEL baseline.

\subsection{RealPatch}\label{sec:realpacth_results}
In this section we include 1) the information to confirm the effect of RealPatch hyperparamaters (further to the \hyperref[sec:ablation]{Ablation Analysis}), 2) additional examples of RealPatch and CycleGAN counterfactuals for both CelebA and Waterbirds datasets, 3) subgroup results for each dataset, 4) examples of matched pairs and achieved matching quality for iWildCam-small, and 5) examples of matched pairs and achieved matching quality for imSitu dataset.

\begin{figure*}[t!]
    \centering
    \includegraphics[scale=0.16]{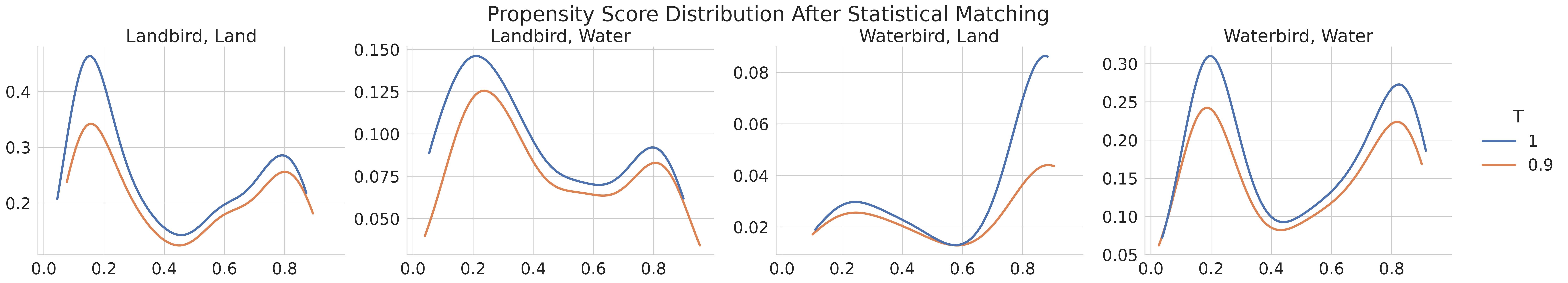}
    \caption{Estimated propensity score distributions on the Waterbirds dataset after matching, shown for each of the four subgroups. We compare the original distribution (blue, $t\!=\!1$) with its scaled version using the selected temperature (orange, $t\!=\!0.9$). Post-matching, the propensity score is approximately bimodal, showing that our procedure is balancing the propensity distribution across subgroups. Decreasing $t$ makes the two modes have more similar values, resulting in a matched dataset with better covariate balance in terms of SMD and VR (Table~\ref{tab:ablation_balance}).}
    \label{fig:ablation_ps_waterbirds}
\end{figure*}

\paragraph{Effect of Temperature on Propensity Score.} For a single run of Waterbirds, in Figure \ref{fig:ablation_ps_waterbirds} we show the estimated propensity score distribution for each of the four subgroups for the dataset obtained after matching $D^{\star}$. Similarly to Figure~\ref{fig:ablation_ps_celeba} we compare the distributions obtained when imposing no temperature scaling ($t\!=\!1$) and when selecting the temperature hyperparameter (here, $t\!=\!0.9$). The figure shows consistent results with what was already observed in CelebA: decreasing $t$ leads to the two modes having more similar values, resulting in matched dataset with better propensity score balance and covariate balance in terms of \textit{SMD} and \textit{VR} (Table~\ref{tab:ablation_balance}).

\begin{figure*}[h!]
     \centering
     \begin{subfigure}[b]{0.49\textwidth}
         \centering
         \includegraphics[scale=0.575]{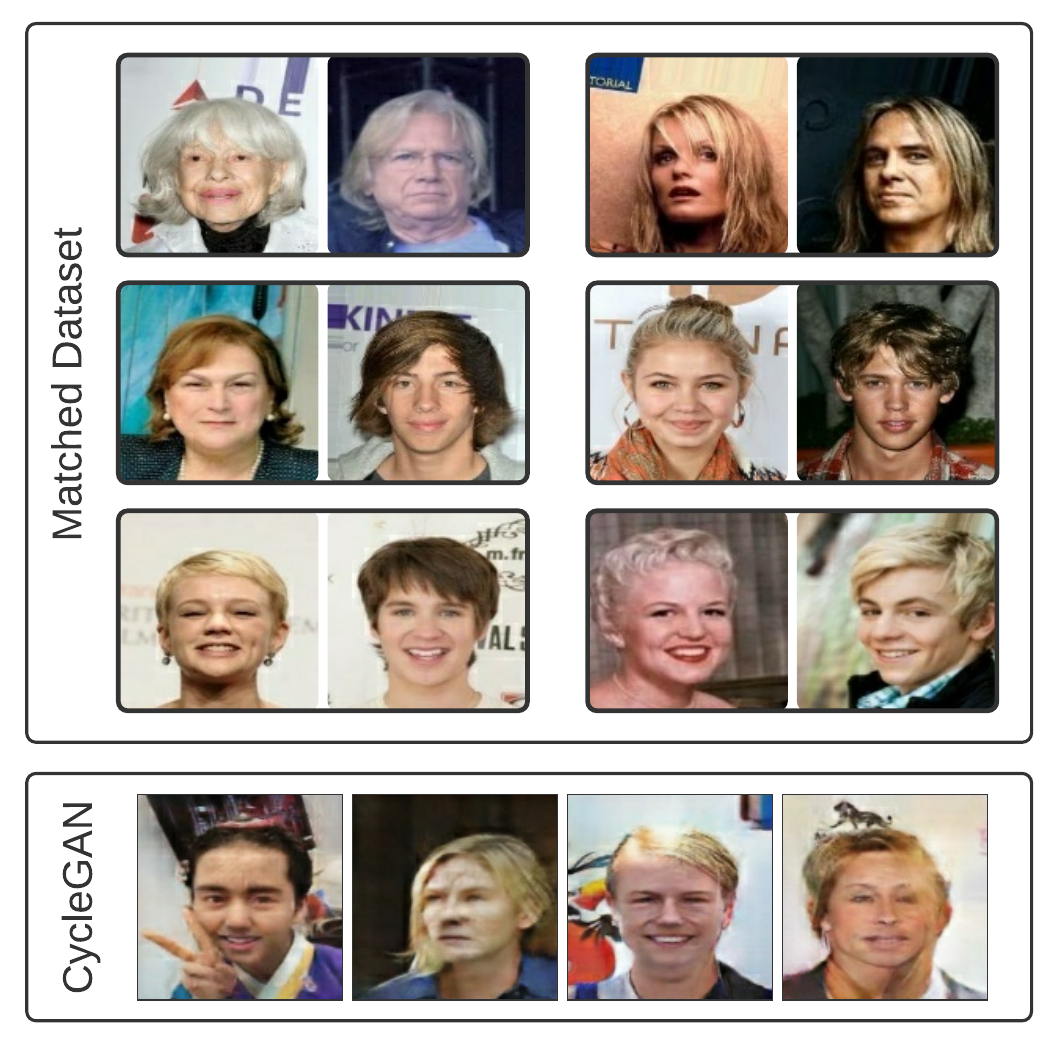}
         \caption{Examples of female images and their male counterfactuals.}
     \end{subfigure}
     \hfill
     \begin{subfigure}[b]{0.49\textwidth}
         \centering
         \includegraphics[scale=0.575]{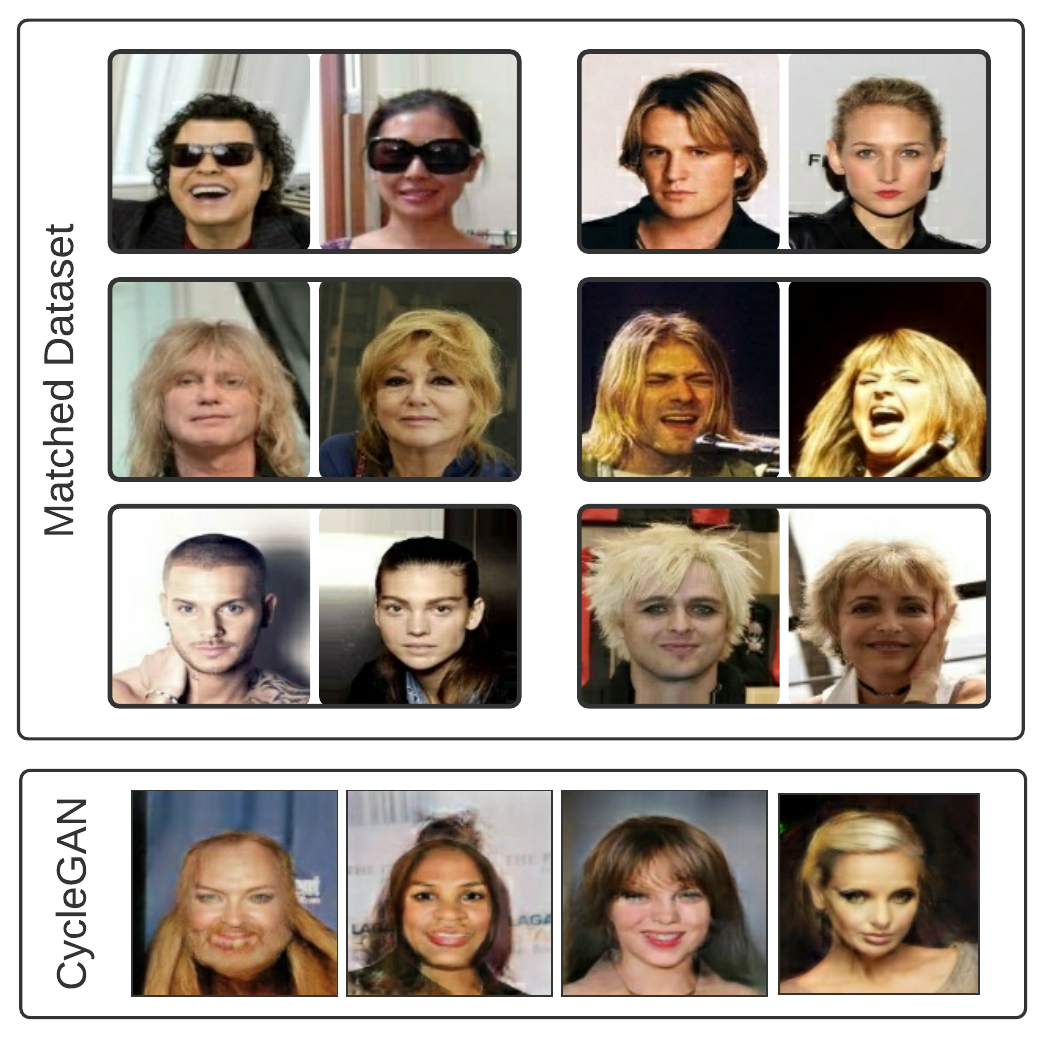}
         \caption{Examples of male images and their female counterfactuals.}
     \end{subfigure}
    \caption{Examples of pairs retrieved using Stage 1 of RealPatch (top); both original and matched images are real samples from the CelebA dataset. We also show CycleGAN synthetic counterfactual results (bottom) on the same attribute-translation task.}
    \label{fig:celeba_example_appendix}
\end{figure*}

\begin{figure*}[t]
     \centering
     \begin{subfigure}[b]{0.49\textwidth}
         \centering
         \includegraphics[scale=0.575]{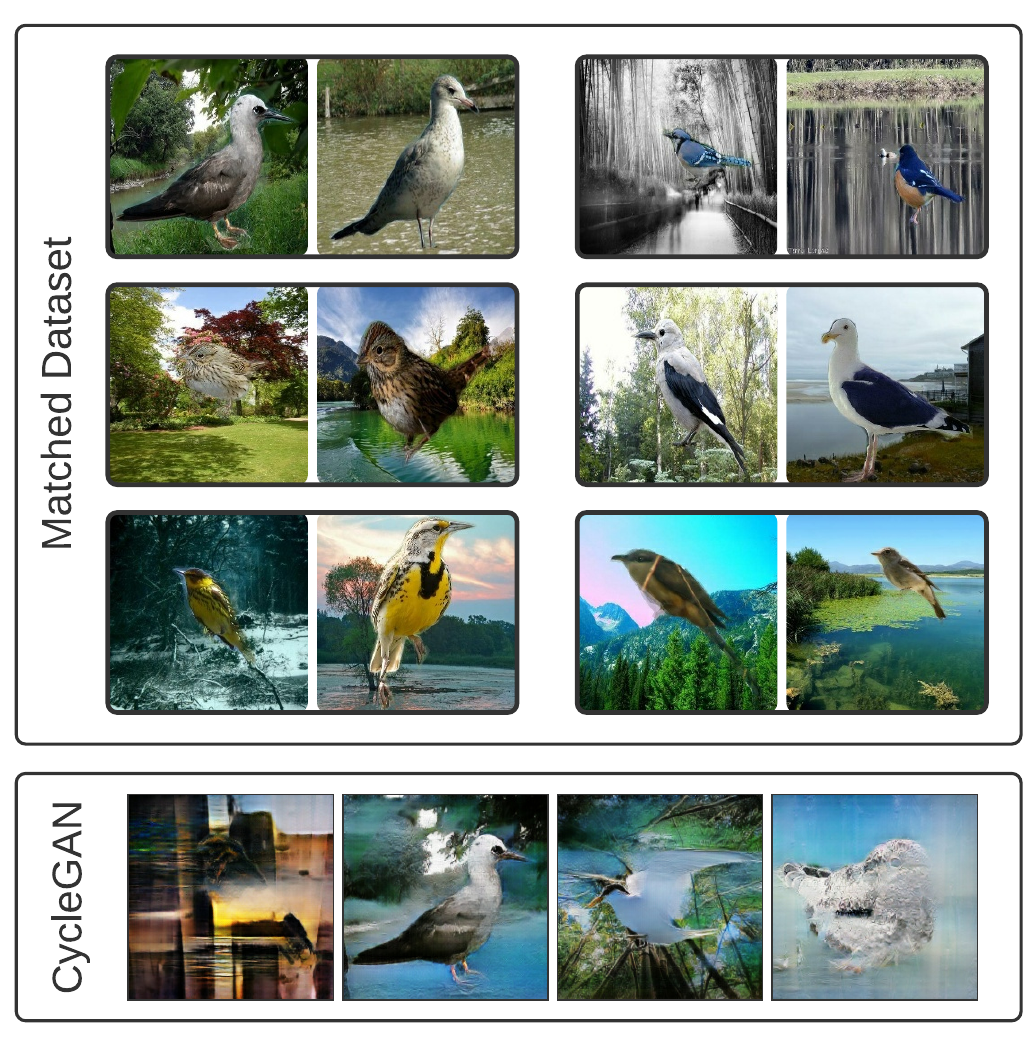}
         \caption{Examples of birds on land and their counterfactuals in water.}
         \label{fig:waterbirds_example_appendix_l_to_w}
     \end{subfigure}
     \hfill
     \begin{subfigure}[b]{0.49\textwidth}
         \centering
         \includegraphics[scale=0.57]{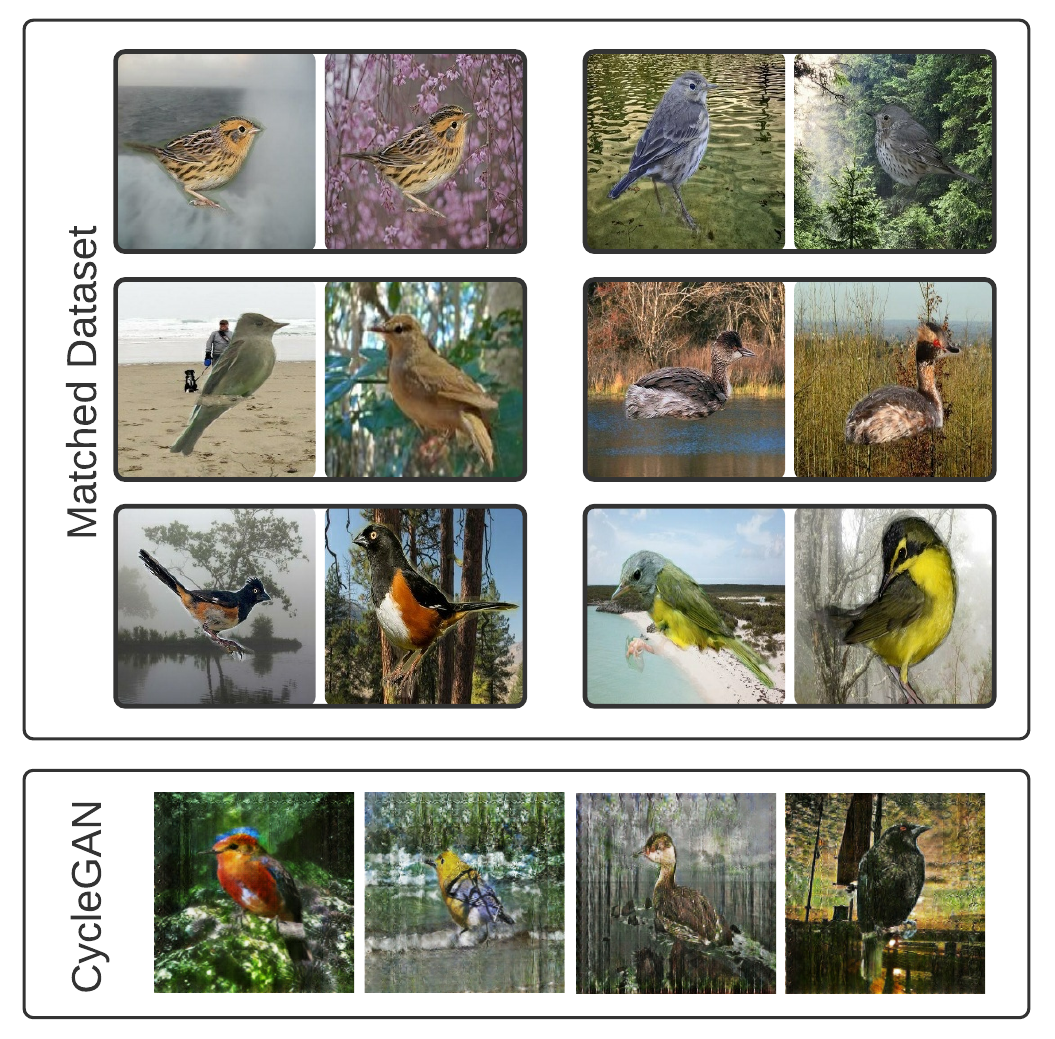}
         \caption{Examples of birds on water and their counterfactuals in land.}
         \label{fig:waterbirds_example_appendix_w_to_l}
     \end{subfigure}
    \caption{Examples of pairs retrieved using Stage 1 of RealPatch (top); both original and matched images are real samples from the Waterbirds dataset. We also show CycleGAN synthetic counterfactual results (bottom) on the same attribute-translation task. CycleGAN often adds artifacts and is frequently unable to recognise birds in the Waterbirds dataset (often removing them when translating from land to water; see left-column \ref{fig:waterbirds_example_appendix_l_to_w}).}
    \label{fig:waterbirds_example_appendix}
\end{figure*}

\paragraph{Additional Counterfactual Examples.} In this section we show additional samples of retrieved matched pairs as well as random synthetic examples generated using CycleGAN. For the CelebA dataset, in Figure~\ref{fig:celeba_example_appendix} we include our results (a) when matching females-to-males and (b) males-to-females. Similarly, for the Waterbirds dataset we include in Figure~\ref{fig:waterbirds_example_appendix} the matched pairs (a) land-to-water and (b) water-to-land. In both datasets we notice that CycleGAN often adds artifacts and is frequently unable to recognise birds in the Waterbirds dataset (often removing them when translating from land to water; see Figure \ref{fig:waterbirds_example_appendix_l_to_w}).

\paragraph{Subgroup results.} Table~\ref{tab:results_subgroup} and Table~\ref{tab:results_subgroup_iwildcam} are an extension of Table~\ref{tab:results_aggregate} and Table~\ref{tab:results_iwildcam} to include the accuracy of all the four subgroups. 
It is worth mentioning that the worst-case accuracy can be observed in different subgroups across the three runs; therefore the average robust accuracy does not necessarily correspond to the average accuracy of one of the four subgroups. 
Although we observe degradation on a subgroup(s) to improve the worst-case in all methods including baselines, our RealPatch makes strikingly better trade-off of the aggregate and robust accuracies than all the baselines. 

\begin{table*}[t!] 
	\begin{center}
		\medskip
		\caption{A comparison between RealPatch and four baselines on two benchmark datasets which includes the subgroup results. This table is an extension of Table~\ref{tab:results_aggregate}. The results shown are the average (standard deviation) performances over three runs.}
		\begin{adjustbox}{width=\textwidth}
			\begin{tabular}{lllllllll}
				\toprule
				\textbf{Dataset} & \textbf{Method} & \makecell{\textbf{Aggregate} $\uparrow$\\ \textbf{Acc. (\%)}} & \makecell{\textbf{Robust} $\uparrow$\\ \textbf{Acc. (\%)}} & \makecell{\textbf{Robust} $\downarrow$ \\ \textbf{Gap (\%)}} & \multicolumn{4}{c}{\makecell{\textbf{Subgroup} $\uparrow$ \quad Y \\ \textbf{Acc. (\%)}   \quad Z}}  \\
				\midrule
				& & & & & Non-Blonde, & Non-Blonde, & Blonde, & Blonde, \\
				& & & & & Female & Male  & Female & Male\\
				\cmidrule{6-9} 
				\textbf{CelebA} & \textbf{ERM} & 89.21 (0.32) & 55.3 (0.65) & 43.48 (0.68) & 80.2 (0.78) & 98.78 (0.11) & 98.07 (0.39) & 55.3 (0.65) \\
			    & \textbf{GDRO} & \textbf{90.47} (7.16) & 63.43 (18.99) & 34.77 (19.65) & 90.03 (10.21) & 92.5 (9.57) & 87.66 (11.07) & 68.75 (26.15) \\
			    & \textbf{SGDRO} &  88.92 (0.18) & 82.96 (1.39) & 7.13 (1.67) & 90.09 (0.31) & 87.67 (0.38) & 88.52 (1.29) & 82.96 (1.39)\\
				& \textbf{CAMEL} & 84.51 (5.59) & 81.48 (3.94) & \textbf{5.09} (0.44) & 85.57 (5.48) & 82.51 (5.26) & 84.15 (2.50) & 81.63 (3.70) \\  
				& \textbf{RealPatch (Our)} &  89.06 (0.13) & \textbf{84.82} (0.85) & 5.19 (0.9) &  90.01 (0.05) & 87.78 (0.14) & 89.52 (0.63) & 84.82 (0.85)\\
				\midrule
				& & & & & Landbird, & Landbird, & Waterbird, & Waterbird, \\
				& & & & & Land & Water  & Land & Water\\
				\cmidrule{6-9} 
				\textbf{Waterbirds} & \textbf{ERM} & 86.36 (0.39) & 66.88 (3.76) & 32.57 (3.95) & 99.45 (0.22) & 76.39 (1.36) & 66.88 (3.76) & 94.95 (0.5) \\
				& \textbf{GDRO} & \textbf{88.26} (0.55) & 81.03 (1.16) & 14.80 (1.15) & 95.83 (0.36) & 81.03 (1.16) & 83.01 (0.7) & 92.2 (0.81) \\
				& \textbf{SGDRO} &  86.85 (1.71) & 83.11 (3.65) & 6.61 (6.01) &  88.53 (4.08) & 85.99 (1.95) & 84.63 (4.81) & 86.19 (1.56)\\
				& \textbf{CAMEL} & 79.0 (14.24) & 76.82 (18.0) & 7.35 (5.66) & 77.17 (17.39) & 82.08 (12.23) & 84.17 (12.86) & 78.85 (19.72) \\
				& \textbf{RealPatch (Our)} &  86.89 (1.34) & \textbf{84.44} (2.53) & \textbf{4.43} (4.48) & 88.03 (3.03) & 86.39 (1.1) & 85.67 (3.54) & 85.93 (0.78)\\
				\bottomrule
			\end{tabular}
		\end{adjustbox}
		\label{tab:results_subgroup}
	\end{center}
\end{table*}

\begin{table*}[t!] 
	\begin{center}
		\medskip
		\caption{A comparison between RealPatch and two baselines on iWildCam-small datasets which includes the subgroup results. This table is an extension of Table\ref{tab:results_iwildcam}. The results shown are the average (standard deviation) performances over three runs.}
		\begin{adjustbox}{width=\textwidth}
			\begin{tabular}{llllllll}
			\toprule
			\textbf{Method} & \makecell{\textbf{Aggregate} $\uparrow$\\ \textbf{Acc. (\%)}} & \makecell{\textbf{Robust} $\uparrow$\\ \textbf{Acc. (\%)}} & \makecell{\textbf{Robust} $\downarrow$ \\ \textbf{Gap (\%)}} & \multicolumn{4}{c}{\makecell{\textbf{Subgroup} $\uparrow$ \quad Y \\ \textbf{Acc. (\%)}   \quad Z}}  \\
			\midrule
			& & & & Meleagris Ocellata, & Meleagris Ocellata, & Crax Rubra, & Crax Rubra, \\
			& & & & 26 & 255  & 26 & 255\\
			\cmidrule{5-8} 
			\textbf{ERM} & \textbf{79.97} (1.18) & 75.43 (3.01) & 19.65 (1.96) & 84.31 (7.33) & 87.07 (2.95) & 92.22 (4.8) & 75.43 (3.01)\\
			\textbf{SGDRO}  & 78.55 (2.45) & 75.5 (3.58) & 14.28 (4.35)  & 85.49 (4.93) & 87.5 (3.06) & 79.25 (2.76) & 75.5 (3.58)\\
			\textbf{RealPatch (Our)} & 79.36 (2.09) & \textbf{76.7} (3.19) & \textbf{11.36} (4.87) & 87.06 (4.8) & 84.58 (2.95) & 80.37 (1.38) & 76.76 (3.26)\\
			\bottomrule
			\end{tabular}
		\end{adjustbox}
		\label{tab:results_subgroup_iwildcam}
	\end{center}
\end{table*}				
				
\begin{figure}[th!]
    \centering
    \includegraphics[scale=0.825]{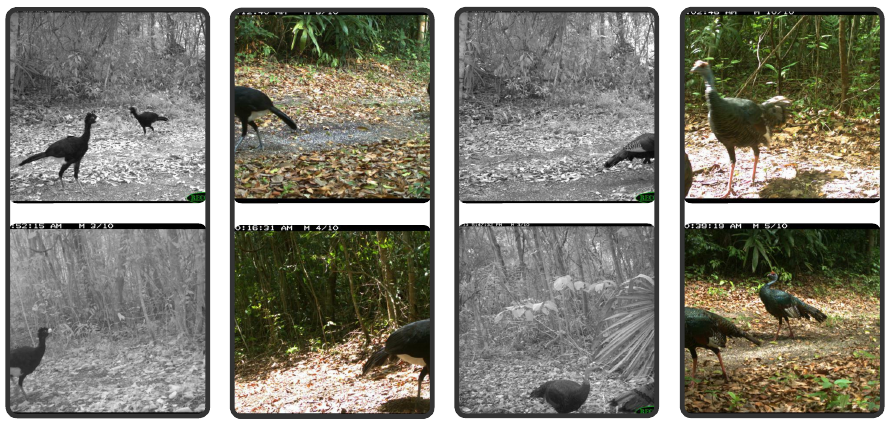}
    \caption{Matched samples on a subset of iWildCam dataset using the spurious attribute camera trap location.}
    \label{fig:iwildcam_matching}
\end{figure}

\paragraph{Additional Results for iWildCam-small.} In Figure~\ref{fig:iwildcam_matching} we show samples of retrieved matched pairs, here we can see how matching is able to preserve the bird species as well as the background colours. 
Similarly to Table~\ref{tab:ablation_balance}, in Table~\ref{tab:ablation_balance_iwildcam} we compare the effect of the main components of our statistical matching stage for iWildCam-small dataset, analysing the effect of temperature scaling and calipers. The selected best configuration for all three runs do not include the usage of std-based caliper, therefore we do not study the effect of removing such component (i.e. setting $\alpha=\infty$). The results are consistent with what observed for CelebA and Waterbirds: the strongest effect is obtained by removing the influence of the fixed caliper, while the impact of temperature scaling is weaker overall.

\begin{table}[t!]
\begin{center}
    \caption{Comparison of the covariate balance in 1) the original dataset $D$, 2) the matched dataset $D^{\star}$ 3) the matched dataset $D^{\star}$ with no temperature scaling 4) $D^{\star}$ with no fixed caliper. The results are reported for a single run per dataset. Our matching procedure can successfully improve the covariate balance in iWildCam-small dataset, with fixed caliper significantly boosting its quality.}
	\medskip
	\scalebox{.85}{
	\begin{tabular}{lllllll}
		\toprule
		 & \multicolumn{3}{c}{\makecell{\textbf{SMD}}} & \multicolumn{3}{c}{\makecell{\textbf{VR}}} \\
        \midrule
		& $\le 0.1$ $\uparrow$ & $(0.1, 0.2)$ $\downarrow$ & $\ge 0.2$ $\downarrow$ & $\le 4/5$ $\downarrow$ & $(4/5, 5/4)$ $\uparrow$ & $\ge 5/4$ $\downarrow$\\
		\cmidrule(lr){2-4} \cmidrule(lr){5-7}
		$D$ & 413 & 354 & 1281  & 612 & 471 & 965\\ 
		$D^{\star}$ (best) & \textbf{1125} & 656 & 267 & 161 & \textbf{1005} & 882\\ 
		$D^{\star}\,\,(t\!=\!1)$ & 753 & 615 & 680 & 191 & 695 & 1162\\
		$D^{\star}\,\,(c\!=\!0)$ & 1037 & 641 & 370 & 331 & 930 & 787\\
		\bottomrule
	\end{tabular}
	}
	\label{tab:ablation_balance_iwildcam}
\end{center}
\end{table}

\begin{figure*}[t]
     \centering
     \begin{subfigure}[b]{0.49\textwidth}
         \centering
         \includegraphics[scale=0.575]{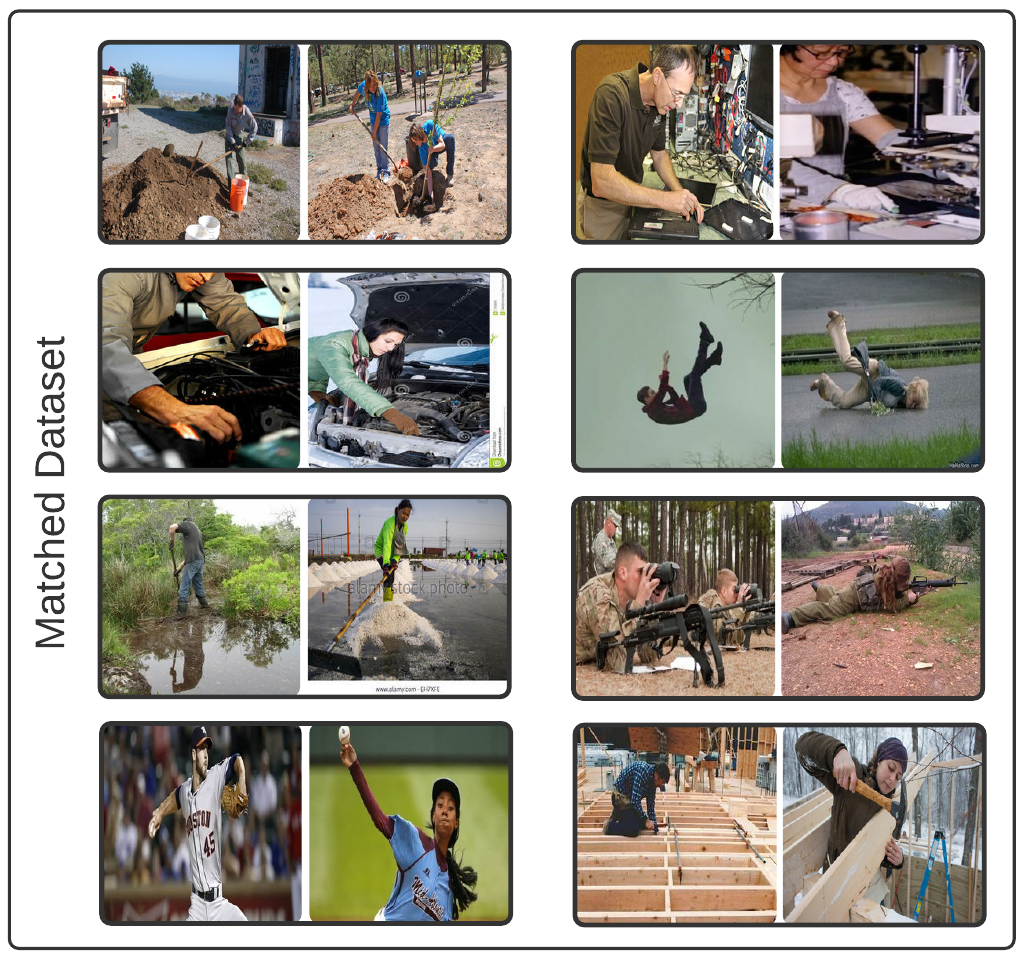}
         \caption{Examples of male images and their female counterfactuals}
     \end{subfigure}
     \hfill
     \begin{subfigure}[b]{0.49\textwidth}
         \centering
         \includegraphics[scale=0.575]{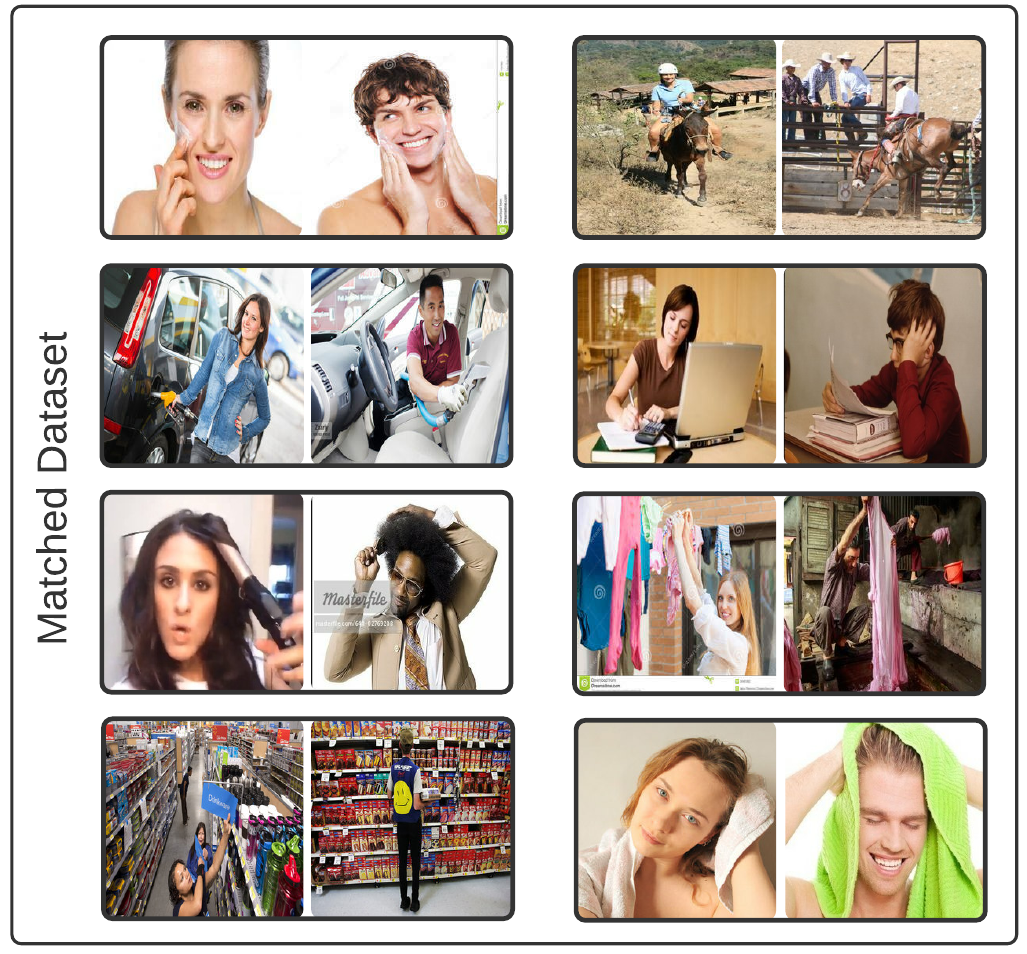}
         \caption{Examples of female images and their male counterfactuals}
     \end{subfigure}
    \caption{Examples of pairs retrieved using using Stage 1 of RealPatch; both original and matched images are real samples from the imSitu dataset. Note that activities are generally preserved across pairs despite not conditioning on the target class during matching.}
    \label{fig:imsitu_example_appendix}
\end{figure*}

\paragraph{Additional Results for imSitu.} In Figures \ref{fig:imsitu_example_appendix} we show examples of matched pairs retrieved using RealPatch on the imSitu dataset. Here, we observe that the activity is generally preserved, though not necessary reflecting an identical \emph{situation} label in the dataset; for example, we have matched images of agents ``pumping'' and ``cleaning'' a car (both related to car maintenance) or agents ``curling'' and ``combing'' hair (both related to hair styling). Additionally, in Table~\ref{tab:matching_balance_imsitu} we show the comparison of the achieved covariates balance imSitu: RealPatch is able to produce a matched dataset with the majority of coviarates perfectly balanced in term of \textit{SMD} and \textit{VR}.

\begin{table}[t!]
	\caption{A comparison of the covariates balance in imSitu, before matching ($D$) and after matching ($D^{\star}$). Our procedure is able to produce a dataset with the majority of covariates perfectly balanced (992 and 1010 out of 1024) in term of \textit{SMD} and \textit{VR}.}
	\begin{center}
	\scalebox{0.85}{
		\medskip
			\begin{tabular}{lllllll}
				\toprule
				 & \multicolumn{3}{c}{\textbf{SMD}} & \multicolumn{3}{c}{\textbf{VR}}\\
				\midrule
            	 & $\le 0.1$ $\uparrow$ & $(0.1, 0.2)$ $\downarrow$ & $\ge 0.2$ $\downarrow$ & $\le 4/5$ $\downarrow$ & $(4/5, 5/4)$ $\uparrow$ & $\ge 5/4$ $\downarrow$\\
				\cmidrule(lr){2-4} 	\cmidrule(lr){5-7}
				$D$ & 327 & 271 & 426 & 272 & 510 & 242\\ 
				$D^{\star}$ & \textbf{992} & 32 & 0 & 4 & \textbf{1010} & 10\\ 
				\bottomrule
			\end{tabular}}
		\label{tab:matching_balance_imsitu}
	\end{center}
\end{table}

\subsection{CAMEL}\label{sec:camel_results}
In Table \ref{tab:results_camel} we report three results for the CAMEL model: (a) the metrics obtained after training the model for full 50 epochs for CelebA (and 500 epochs for Waterbirds) as per \cite{goel2020model}; (b) the results from the epoch where the model achieved the best robust metric on the validation set; in accordance with RealPatch, we report the average (standard deviation) across three repeats over different data splits for both (a) and (b) results; (c) the results from Table 2 in \cite{goel2020model} are also included since the authors have a different setup, namely they keep the default train/validation/test split while \emph{changing the random seed to initialise the model}.
We include both settings (a) and (b) since the exact procedure in \cite{goel2020model} is somewhat unclear; we use authors' implementation of CAMEL to produce them. The results in Table~\ref{tab:results_aggregate} show the output of the method described in (b) as it appears to be the closest comparison.

\begin{table*}[t!]
	\begin{center}
	\caption{Three different results for the CAMEL model: 1) metrics obtained at the last epoch after training the model; 2) results from the epoch where the model achieved the best robust gap on the validation set; 3) the results included in Table 2 in \cite{goel2020model}.}
		\medskip
		\begin{adjustbox}{width=0.8\textwidth}
    		\begin{tabular}{llllll}
    			\toprule
    			\textbf{Dataset} & \textbf{Method} & \makecell{\textbf{Aggregate} $\uparrow$\\ \textbf{Acc. (\%)}} & \makecell{\textbf{Robust} $\uparrow$ \\ \textbf{Acc. (\%)}} & \makecell{\textbf{Robust} $\downarrow$\\ \textbf{Gap (\%)}} \\
    			\midrule
    			\textbf{CelebA} & \textbf{CAMEL} (re-run epoch 50) & 96.6 (0.51) & 57.96 (3.55) & 40.12 (4.18) \\ 
    			& \textbf{CAMEL} (re-run SGDRO) & 84.51 (5.59) & 81.48 (3.94) & 5.09 (0.44) \\   
    			& \textbf{CAMEL} \cite{goel2020model}, Table 2 & 92.90 (0.35) & 83.90 (1.31) & - \\  
    			\midrule
    			\textbf{Waterbirds} & \textbf{CAMEL} (re-run epoch 500) & 89.63 (7.84) & 68.12 (6.93) & 29.59 (3.91)\\
    			& \textbf{CAMEL} (re-run SGDRO) & 79.0 (14.24) & 76.82 (18.0) & 7.35 (5.66) \\  
    			& \textbf{CAMEL} \cite{goel2020model}, Table 2 & 90.89 (0.87) & 89.12 (0.36) & - \\
    			\bottomrule
    		\end{tabular}
		\end{adjustbox}
		\label{tab:results_camel}
	\end{center}
\end{table*}

\end{document}